\documentclass[conference,10pt]{IEEEtran}
\IEEEoverridecommandlockouts

\usepackage{cite}
\usepackage{amsmath,amssymb,amsfonts}

\usepackage{algorithm}
\usepackage{algpseudocode}
\usepackage{graphicx}
\graphicspath{{./}{figs/}}
\usepackage[T1]{fontenc}
\usepackage{textcomp}
\usepackage{xcolor}
\usepackage{booktabs}
\usepackage{multirow}
\usepackage{array}
\usepackage{url}
\usepackage{hyperref}
\usepackage{pgfplots}
\usepackage{tikz}
\usetikzlibrary{shapes,arrows,positioning,fit,backgrounds,calc,
                decorations.pathreplacing,matrix}
\pgfplotsset{compat=1.17}

\newcommand{\Fhat}{\hat{F}}
\newcommand{\sema}{s_{\mathrm{ema}}}
\newcommand{\hema}{h_{\mathrm{ema}}}
\newcommand{\snew}{s_{\mathrm{new}}}
\newcommand{\hnew}{h_{\mathrm{new}}}

\begin{document}

\title{CAFIN: CUDA-accelerated Adaptive Fisher Information Matrix Network Optimization via GPU-Native Kernel Fusion}

\author{
\begin{minipage}[t]{0.48\linewidth}
\centering
\textbf{Maryam Arshad}\\
SEECS, National University of Sciences and Technology (NUST)\\
Islamabad, Pakistan\\
marshad.msee23seecs@seecs.edu.pk
\end{minipage}
\hfill
\begin{minipage}[t]{0.48\linewidth}
\centering
\textbf{Muhammad Khuram Shahzad}\\
SEECS, National University of Sciences and Technology (NUST)\\
Islamabad, Pakistan\\
mkhurram.shahzad@seecs.edu.pk
\end{minipage}
}

\maketitle
\begin{abstract}
Second-order optimizers can improve the convergence behavior of deep neural networks by incorporating curvature information, but their practical use is often limited by the computational overhead of optimizer-side GPU operations. AdaFisher reduces the cost of Fisher Information Matrix (FIM) approximation through a diagonal block-Kronecker representation and exponential moving averages; however, its PyTorch implementation still relies on multiple sequential tensor operations that introduce repeated CUDA kernel launches, intermediate tensor materialization, and redundant global-memory accesses. This paper presents CAFIN, a GPU-native CUDA kernel fusion approach for improving the execution efficiency of AdaFisher without changing its underlying optimization formulation. The proposed approach progressively fuses per-layer preconditioner construction, exponential moving average updates, gradient preconditioning, momentum accumulation, weight decay, and parameter updates into increasingly integrated CUDA kernels. Shared-memory reuse, register-resident computation, and in-place operations are further employed to reduce redundant memory accesses and eliminate unnecessary intermediate tensor materialization. The proposed implementation is evaluated using ResNet-18 on CIFAR-10, CIFAR-100, and Tiny ImageNet across NVIDIA GTX 1660 Ti and RTX 3060 Laptop GPUs. Experimental results show up to a 78.4\% reduction in isolated FIM construction latency and a 2.1--26.9\% reduction in optimizer-step latency, depending on the execution configuration and hardware. The fused implementation also eliminates approximately 18.91~MB of intermediate GPU memory allocation per optimizer step. Numerical validation shows that the fused implementation remains within the predefined FP32 numerical tolerance of the reference implementation, while convergence accuracy remains comparable across the evaluated datasets and random seeds. These results demonstrate that execution-level GPU optimization through kernel fusion, memory reuse, and elimination of intermediate tensor materialization can substantially improve the practical efficiency of second-order optimization while preserving the learning behavior of AdaFisher.
\end{abstract}


\section{Introduction}
\label{sec:intro_related}

\subsection{Introduction}

Deep neural networks are commonly trained using first-order optimization methods
such as Stochastic Gradient Descent (SGD)~\cite{robbins1951},
Adam~\cite{kingma2014adam}, and AdamW~\cite{loshchilov2019decoupled}. These methods
are efficient and widely used, but rely solely on gradient information and do not
account for the local curvature of the loss landscape, which can slow convergence
in ill-conditioned regions.

Second-order optimization methods incorporate curvature information, typically via
the Hessian or Fisher Information Matrix (FIM), to adapt parameter updates to the
local geometry of the loss function. While these methods can improve convergence
speed and stability, direct computation or inversion of full curvature matrices is
intractable for modern networks due to quadratic scaling with the number of
parameters.

Structured approximations, such as K-FAC~\cite{martens2015kfac},
Shampoo~\cite{gupta2018shampoo}, AdaHessian~\cite{yao2021adahessian},
Sophia~\cite{liu2023sophia}, and AdaFisher~\cite{adafisher2024}, reduce
computational and memory complexity via block-diagonal, Kronecker-factored, or
diagonal approximations, retaining many benefits of curvature-aware training while
remaining tractable.

Recent advances in deep learning have also highlighted the importance of
efficient neural architectures and computationally optimized training
strategies. Modern AI applications require models that maintain predictive
performance while reducing computational overhead and resource demands.
Studies on efficient neural architectures and attention-based learning
frameworks demonstrate the increasing need for optimization strategies that
balance accuracy and computational efficiency \cite{israr2023attention}.

These algorithmic breakthroughs still do not help get past the challenges introduced by GPU training, particularly the issue of available memory.
Optimizer-side tensor operations in execution. Frameworks such as
PyTorch~\cite{pytorch} create a tensor operation and execute it directly on a CUDA, starting with each individual operation.
kernel. This is a sequential transformations, will generate multiple kernel launches, intermediate results.
Rapidly allocates memory, and accesses the memory repeatedly, accounting for most of the execution time for
Memory-bound workloads with a focus on consumer GPUs.

The author (AdaFisher~\cite{adafisher2024}) recovers per-layer FIMs from EMA statistics of
Usually, six CUDA kernels are needed per layer for activations and gradients. While
This sequential execution has high overhead as it is mathematically simple.
This raises the central questions in this piece of work: can the AdaFisher optimizer
Comprising the work, pipeline be fused to be closer to the math operations of the GPU.
formulation?

Based on this, we suggest a three-stage progressive GPU native kernel fusion framework since the traditional CPU-FPGA based method is heavily restricted in hardware support.
Step (i)~fused preconditioner construction, (ii)~fused EMA-plus-reconstruction, and (iii)~ Parietal fusion are areas in which further research is needed.
(iii) An optimizer step kernel allowing for complete integration of the EMA updates, FIM
Recovery, momentum, weight decay and parameter updates. Experiments on
ResNet-18 achieves significant results on CIFAR-10, CIFAR-100 and Tiny ImageNet.
Increased convergence and decreased latency and memory use across
multiple random seeds. In particular, the fully-fused optimizer step saves about
Solving practical problems is the reason of running 18 hours.No less than 10 hours or a bit less than 1.13~hours was reduced during an 18 hours training run, a practically meaningful reduction for.
Experimental work using consumer-grade hardware, with iterative experiments. This study aims to focus on the role of this work in Morocco:
Why systems optimization is essential in effectively achieving practical benefits of
second-order neural-network training.

This paper makes the following main contributions:
\begin{itemize}
    \item \textbf{Progressive GPU-native framework for fusion:} For AdaFisher, we develop a three-stage kernel-fusion framework, which takes advantage of the CUDA platform and gradually integrates the construction of the preconditioner, the EMA update and the final step of the optimizer. This design not only uses the original 6 per Layer CUDA kernel launch to be streamlined to one, but it also decouples the performance impact each level in the fusion hierarchy creates.

    Shared-memory and register-resident optimizer implementation: We implement a fully fused CUDA version by reusing the EMA values inside the shared memory and compute and consume the FIM elements inside the register. This helps prevent repeated DRAM access and intermediate FIM materialization that would reduce the usage of all involved intermediate GPU allocations by around 18.91 MB per optimizer step.

    We perform empirical evaluation of the CAFIN on ResNet-18 for three popular datasets CIFAR-10, CIFAR-100, and Tiny ImageNet, on two popular GPUs NVIDIA Turing and Ampere. The evaluation confirms up to 78.4

    As well as numerical correctness testing, the roofline model, Amdahl's model, memory measurement analysis, and statistical evaluation of all these analyses, we are investigating the pros and cons of employing fusion -- not only whether it leads to better performance on AdaFisher, but also why this is the case, and how these gains translate into real-world training time savings.
\end{itemize}

The rest of this paper is organized as follows. In Section II, theadaFisher preconditioning formervation is described, and its bottle necks while running it on the GPU are identified. In Section III, we introduce the proposed CAFIN methodology and our progressive CUDA kernel-fusion strategies. Section IV presents the experimental evaluation, covering how the experiments were set up. Section V provides a statistical and performance analysis of what we measured, why it matters, how sure we are about it, etc. To wrap up, Section VI provides a summary of the main conclusions and suggests areas for future research.

\subsection{Related Work}

\textbf{Second-order optimization.}
We consider the use of the FIM for the natural gradient method \cite{amari1998natural} here.
Some of the parameter-invariant updates are available, and it is impractical to store and inverse the whole FIM.
Structured approximations like
K-FAC~\cite{martens2015kfac,ba2017distributed,osawa2019large},
Shampoo~\cite{gupta2018shampoo}, AdaHessian~\cite{yao2021adahessian},
Sophia~\cite{liu2023sophia} and AdaFisher~ \cite{adafisher2024} do both aim to reduce.
We are designed to be cost effective and still capture the curvature information. All of these have GPU execution.
challenges due to sequential tensor operations.

Combine kernels and optimize GPU.
communication steps, and allows for dense data matrices or vectors.Combining GPU operations reduces the overhead incurred from launching kernels, and avoids intermediate (cross-machine)
Vectors, matrices, and tensors are materialized.This materializes vectors, matrices and tensors, and enhances data
locality~\cite{wahib2014scalable,fauzia2015characterizing}.

New algorithms and practices!Efficient deep learning practices!
In practical, real-world applications AI implementations have grown in significance.
deployment. Recent deep learning based systems show that reducing data storage costs for cloud computing is achievable.
For computational cost and the performance of the models to be compatible, it is necessary to:
real-world applications. As a result, optimization strategies occur in line with the developments.
that not only enhance the design of the model but also the underlying execution.
To perform deep learning workloads with efficiency \cite{arshad2024deep}.

Compiler frameworks
As described above, there are tools available, including TVM\cite{chen2018tvm} and XLA, as well as TorchInductor\cite{ansel2024pytorch}.
Performs automatic operator fusion, but optimizers keep states;
and make in-place updates to changing groups of parameters, which is more involved because they are dynamic.
automatic fusion and inspires to write CUDA kernels by hand, with or without NVIDIA's AdopTCUDA~\cite{nvidia2023cuda} tools.
Write exploits using Python with the help of Triton~\cite{tillet2019triton}.
It has been adopted for implementing fused attention for tile-based GPU kernels.
operators~\cite{dao2022flashattention,dao2023flashattention2}, but requires
explicit tiling decision that is not justifiable when performing a rank-one outer-product
Continuous EMA States for workloads. CUDA Graphs~\cite{nvidia2019cudagraphs}
forces the watertightness of the kernel-launch to be proved by taking and replaying a fixed sequence of
do not remove the intermediate tensor materialization or make it better, but;but, do not remove intermediate tensor materialization or make it better;
reuse of data in a kernel. NVIDIA Apex~\cite{apex2018} and
DeepSpeed~\cite{rasley2020deepspeed} give fused implementations for first order
A similar facility is provided for second-order methods that is missing for optimizers. Operation-specific
Some of the latest techniques for fusion are FlashAttention~\cite{dao2022flashattention,dao2023flashattention2}.
and FastKron~\cite{fastkron2024} demonstrate a considerable speedup on memory heavy applications.
workloads. Fused primitives are also provided for convolution in cuDNN~\cite{nvidia2023cuda}.
and normalization, but doesn't expose optimizer level fusion apis.

\textbf{Optimizer fusion challenges.}
Optimizers do in-place Updates, persistent states, and iterates
Because this process is dynamic in nature and the technologies are applied at a level of parameters as groups, it becomes more complicated to achieve automatic fusion. In AdaFisher,
Construction and updates of EMA statistics have been enforced in various parameter groups, as reconstruction of EMA is performed across the groups.
Kronecker-structured preconditioners that use more kernels which are hard to solve for example:
to collapse automatically.

\textbf{Positioning.}
In contrast to previous work that put forward novel curvature approximations or optimizer algorithms,
This paper is specifically devoted to using progressive optimization to optimize the AdaFisher system.
kernel fusion. The approach maintains optimizer equations and semantics for training,
Complementing algorithmic development and being future integratable with curvature.
approximation methods.

\section{AdaFisher Preconditioning}
\label{sec:background}

For a given layer $\ell$ with weight matrix
$\theta_\ell \in \mathbb{R}^{d_{\text{out}} \times d_{\text{in}}}$,
AdaFisher approximates the Fisher block as
$F_\ell \approx A_\ell \otimes G_\ell$,
where $A_\ell$ and $G_\ell$ represent the activation and pre-activation gradient
covariance matrices, respectively. To reduce computational cost, only their diagonal
components are maintained:
\begin{equation}
\label{eq:diag_statistics}
  \sema = \mathrm{diag}(A_\ell) \in \mathbb{R}^{d_{\text{in}}},
  \quad
  \hema = \mathrm{diag}(G_\ell) \in \mathbb{R}^{d_{\text{out}}}.
\end{equation}

The diagonal statistics defined in
Equation~\eqref{eq:diag_statistics} provide compact representations
of the activation and pre-activation-gradient curvature factors.

The layer preconditioner is reconstructed element-wise as:
\begin{equation}
  \label{eq:fim}
  \Fhat_{\ell,ij} = \hema[i] \cdot \sema[j] + \lambda,
\end{equation}
so that gradients can be scaled directly without requiring an explicit matrix
inversion. Equation~\eqref{eq:fim} is the central computation targeted by all
fusion stages in Section~\ref{sec:methodology}.

Per-layer statistics are collected using forward and backward
hooks~\cite{adafisher2024}. For a mini-batch of activations
$\{a^{(b)}\}_{b=1}^{B}$ and pre-activation gradients
$\{\delta^{(b)}\}_{b=1}^{B}$, the statistics are computed as:
\begin{align}
\snew &= \frac{1}{B} \sum_{b=1}^{B} a^{(b)} \odot a^{(b)},
\label{eq:snew}\\
\hnew &= \frac{1}{B} \sum_{b=1}^{B} \delta^{(b)} \odot \delta^{(b)}.
\label{eq:hnew}
\end{align}

Equations~\eqref{eq:snew} and \eqref{eq:hnew} compute the
mini-batch activation and pre-activation-gradient statistics,
respectively.

where $\odot$ denotes the Hadamard (element-wise) product. These statistics are accumulated using exponential moving averages (EMA):
\begin{align}
  \sema &\leftarrow \gamma_1 \sema + \gamma_2 \snew, \\
  \label{eq:ema_s}\\
  \hema &\leftarrow \gamma_1 \hema + \gamma_2 \hnew,
  \label{eq:ema_h}
\end{align}
Equations~\eqref{eq:ema_s} and \eqref{eq:ema_h} update the
running activation and gradient statistics through exponential
moving averages.
For all reported experiments, $\gamma=0.8$ is the base EMA decay rate (Table~\ref{tab:hyp}),
giving $\gamma_1=0.08$ and $\gamma_2=0.008$ for all reported experiments.

The gradient is then preconditioned as
$\tilde{G}_\ell = G_\ell \oslash \Fhat_\ell$,
followed by momentum accumulation and parameter update.

In the reference PyTorch implementation, EMA updates and preconditioner
reconstruction are executed via six independent CUDA kernels per layer per
optimizer step: two kernels for EMA updates and four kernels for preconditioner
reconstruction and gradient scaling. Each kernel writes intermediate tensors to
DRAM, generating repeated memory traffic and kernel launch overhead.

For example, for layer4.1 of ResNet-18 ($d_{\text{out}} = d_{\text{in}} = 512$),
the combined intermediate tensor allocations for FIM reconstruction reach
approximately 28.3\,MB per optimizer step. Across all ten ResNet-18 layers,
the total intermediate allocation reaches roughly 37.8\,MB per optimizer step.

Roofline analysis illustrates the memory-bound nature of this workload. Each
output element $\Fhat_{ij}$ requires roughly two FLOPs (one multiplication and one
addition). Assuming $\hema[i]$ is broadcast across an output row and amortized over
$d_{\text{in}}$ elements, the dominant per-element memory traffic includes one
4-byte read of $\sema[j]$ and one 4-byte output write:
\begin{equation}
\label{eq:baseline_ai}
  \mathrm{AI} \approx \frac{2}{8} = 0.25~\text{FLOP/byte}.
\end{equation}
Equation~\eqref{eq:baseline_ai} therefore characterizes the
baseline FIM reconstruction as a low-arithmetic-intensity workload.
With shared-memory reuse of the EMA vectors, the fused implementation increases
arithmetic intensity to approximately 0.5~FLOP/byte. On the GTX~1660~Ti with peak
memory bandwidth of 288.0~GB/s and peak FP32 throughput of 5.0~TFLOPS, the
roofline ridge point is:
\begin{equation}
\label{eq:ridge_point}
  \frac{5000}{288} \approx 17.36~\text{FLOP/byte}.
\end{equation}

Comparing the arithmetic intensity in
Equation~\eqref{eq:baseline_ai} with the ridge point in
Equation~\eqref{eq:ridge_point} confirms that FIM reconstruction
operates well within the memory-bound regime.
Section~\ref{sec:methodology}.

\section{Methodology}
\label{sec:methodology}

\begin{figure*}[t]
  \centering
  \includegraphics[width=\textwidth]{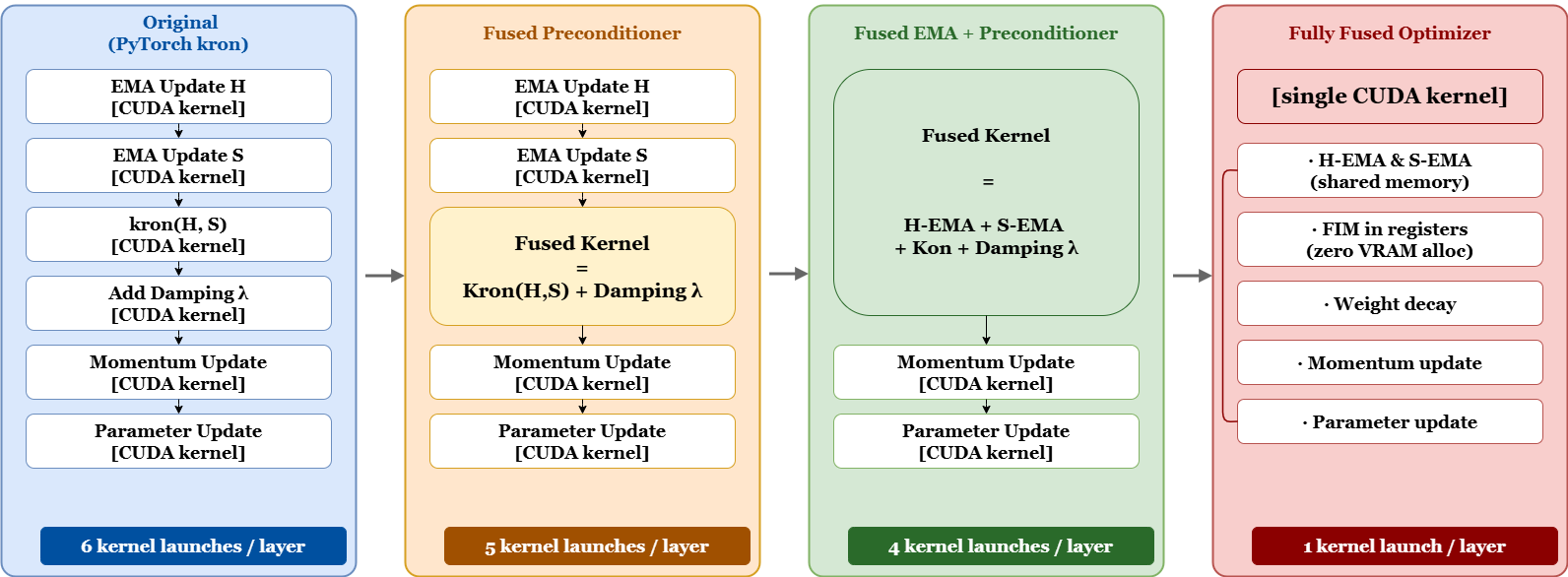}
  \caption{Progressive kernel fusion in AdaFisher showing reduction of per-layer CUDA kernel launches. From left to right: original (6 kernels), fused preconditioner (5 kernels), fused EMA + preconditioner (4 kernels), and fully fused optimizer (1 kernel). The fusion reduces kernel dispatch overhead, eliminates intermediate tensor allocations, and improves GPU execution efficiency.}
  \label{fig:pipeline}
\end{figure*}

Figure~\ref{fig:pipeline} illustrates the progressive reduction in per-layer CUDA
kernel launches across the execution stages.

This work accelerates the AdaFisher optimizer using progressive CUDA kernel
fusion. In the original implementation, the layer-wise Fisher Information Matrix
(FIM) is reconstructed through multiple PyTorch tensor operations, where each
operation launches a separate CUDA kernel and creates temporary tensors in GPU
memory.In our approach, such operations are integrated and executed in a more and more fused manner.
leverages the stages for kernel launch overhead reduction, temporal memory reduction and
improve the efficiency of the use of memory without altering the mathematical characteristics of
the original optimizer.

Optimisation pipeline has three stages:
(i)~fused preconditioner construction,
The (ii) ~ fused exponential moving average (EMA) / preconditioner construction,
(iii) a fully fused optimizer step fusing EMA updates, FIM reconstruction,
The updates to the parameters within the same kernel as well as the accumulation of momenta.

This is important because of the progressive design. First, it isolates the
performance impact of each optimization stage, allowing the contribution of kernel
The evaluation of the fusion, memory reduction and reusing shared-memory is being done separately.
Second, the composition is simplified by gradual fusion, and correctness is facilitated when checking for the composition is preserved.
During: for each stage before numerical behavior of the original AdaFisher-Sort implementation calculation.Before: During numerical behavior of the original AdaFisher-Sort implementation calculation.
Adding the complexity of the fusion. This staged approach also offers an opportunity to show that one's values and ethics support the process.
How each 'bottleneck' contributes to the overall 'bottleneck' process.
optimizer overhead.
\subsection{Baseline AdaFisher Execution}
\label{subsec:baseline}

In the original AdaFisher implementation, the layer-wise preconditioner is
reconstructed as in Equation~\eqref{eq:fim}:
\begin{equation}
  \Fhat_\ell = \hema \, \sema^\top + \lambda I,
  \label{eq:baseline_prec}
\end{equation}
where $\hema$ and $\sema$ denote the running exponential moving averages of the
activation and gradient statistics, respectively, and $\lambda$ is a damping term.

The baseline implementation computes this expression using multiple PyTorch tensor
operations such as \texttt{unsqueeze}, \texttt{kron}, \texttt{transpose}, and
\texttt{add}. Each operation launches a separate CUDA kernel. Temporary tensors
are repeatedly created in GPU memory and then reused by later operations, including
the full $d_{\text{out}} \times d_{\text{in}}$ FIM matrix.

Algorithm~\ref{alg:baseline} summarizes the baseline AdaFisher optimizer pipeline.

\begin{algorithm}[t]
\caption{Reference Baseline AdaFisher Optimizer}
\label{alg:baseline}
\begin{algorithmic}[1]

\Require $\hema, \sema, \hnew, \snew, \nabla\theta, \theta, m,$
         $\gamma_1, \gamma_2, \beta, \eta, \lambda, \omega, b_c,$
         $d_{\mathrm{out}}, d_{\mathrm{in}}$

\Ensure Updated $\theta$, $m$, $\hema$, and $\sema$

\State $\hema \gets \gamma_1\hema + \gamma_2\hnew$
\State $\sema \gets \gamma_1\sema + \gamma_2\snew$

\For{$i = 0$ to $d_{\mathrm{out}}-1$}
    \For{$j = 0$ to $d_{\mathrm{in}}-1$}
        \State $\hat{F}_{ij} \gets
        \hema[i]\sema[j] + \lambda$
    \EndFor
\EndFor

\State $g \gets \nabla\theta + \omega\theta$
\State $m \gets \beta m + (1-\beta)g$
\State $\theta \gets
\theta - \frac{\eta}{b_c}
\left(m \oslash \hat{F}\right)$

\end{algorithmic}
\end{algorithm}

\subsection{Fused Preconditioner Construction}
\label{subsec:fusedop}

The baseline AdaFisher implementation spends a significant portion of execution
time launching multiple CUDA kernels and repeatedly accessing temporary tensors
from GPU memory. Since the FIM reconstruction (Equation~\eqref{eq:fim}) has a
rank-one outer-product structure, each output element can be computed independently
using only one activation EMA value and one gradient EMA value.

The first optimization stage replaces the sequence of PyTorch tensor operations
with a single CUDA kernel, \texttt{fim\_outer\_add}:
\begin{equation}
  \text{out}[i,j] = \hema[i]\cdot\sema[j]+\lambda.
  \label{eq:fused_op}
\end{equation}

The fused CUDA kernel evaluates the element-wise preconditioner
defined in Equation~\eqref{eq:fused_op} directly, without
materializing the intermediate Kronecker-product tensors.
Compared with the dense preconditioner construction in
Algorithm~\ref{alg:baseline}, Algorithm~\ref{alg:fused_precond}
computes each preconditioner element directly within a single
fused CUDA kernel, avoiding the intermediate tensor
materializations required by the baseline implementation, reducing per-layer
kernel launches from 6 to 5 and saving up to 18.91~MB of
intermediate GPU memory per optimizer step.

Algorithm~\ref{alg:fused_precond} summarizes the proposed
Stage~1 fused preconditioner construction.
\begin{algorithm}[t]
\caption{CAFIN Stage 1: Fused Preconditioner Construction}
\label{alg:fused_precond}
\begin{algorithmic}[1]

\Require $\hema, \sema, \lambda,
d_{\mathrm{out}}, d_{\mathrm{in}}$

\Ensure Preconditioner $\hat{F}$

\ForAll{$(i,j)$ such that
$0 \leq i < d_{\mathrm{out}}$ and
$0 \leq j < d_{\mathrm{in}}$ \textbf{in parallel}}

    \State $\hat{F}_{ij}
    \gets \hema[i]\sema[j] + \lambda$

\EndFor

\end{algorithmic}
\end{algorithm}

For reference, we also evaluate an \emph{algebraic simplification} variant of
the baseline, which rewrites Equation~\eqref{eq:baseline_prec} directly as an
element-wise product followed by a scalar add---replacing \texttt{kron} and
\texttt{unsqueeze} chains with a single \texttt{outer} call and \texttt{add}
in PyTorch. This eliminates redundant shape manipulations without requiring a
custom CUDA kernel, and serves as an intermediate reference point between the
eager baseline and the fully fused implementation. The algebraic variant still
launches multiple kernels (one per PyTorch op) and materializes the full FIM
matrix in global memory; it is not a fusion strategy, but rather a cleaner
reimplementation of the same computation.

\subsection{Fused EMA Update and Preconditioner Reconstruction}
\label{subsec:ema_integration}

The second optimization stage combines EMA updates and preconditioner
reconstruction within the same CUDA execution pipeline. Instead of executing the
EMA updates (Algorithm~\ref{alg:baseline}, lines 1--2) as separate kernel
launches before the outer-product kernel, this stage fuses them into a single
kernel that reads the raw statistics $\hnew$ and $\snew$, updates the EMA
accumulators in shared memory, and immediately computes the preconditioner
values. This eliminates an additional two kernel launches per layer, reducing the
total to 4 launches per layer per optimizer step. The fused pipeline also avoids
writing the updated EMA vectors back to global memory before they are consumed
by the outer-product computation.

Algorithm~\ref{alg:fused_ema_precond} summarizes Stage~2,
in which the EMA updates and preconditioner reconstruction
are combined within the same fused execution path.

\begin{algorithm}[t]
\caption{CAFIN Stage 2: Fused EMA and Preconditioner Construction}
\label{alg:fused_ema_precond}
\begin{algorithmic}[1]

\Require $\hema, \sema, \hnew, \snew,
\gamma_1, \gamma_2, \lambda,
d_{\mathrm{out}}, d_{\mathrm{in}}$

\Ensure Updated $\hema$, $\sema$, and preconditioner $\hat{F}$

\State $\hema^{+}
\gets \gamma_1\hema + \gamma_2\hnew$

\State $\sema^{+}
\gets \gamma_1\sema + \gamma_2\snew$

\State Cooperatively cache the required
$\hema^{+}$ and $\sema^{+}$ values in shared memory

\State \Call{SyncThreads}{}

\ForAll{$(i,j)$ such that
$0 \leq i < d_{\mathrm{out}}$ and
$0 \leq j < d_{\mathrm{in}}$ \textbf{in parallel}}

    \State $\hat{F}_{ij}
    \gets \hema^{+}[i]\sema^{+}[j] + \lambda$

\EndFor

\State $\hema \gets \hema^{+}$
\State $\sema \gets \sema^{+}$

\end{algorithmic}
\end{algorithm}

On Tiny ImageNet, this stage exhibits a minor latency regression ($-3.5\%$
relative to the Fused Preconditioner stage) for larger layers. Profiling reveals
that for layers with $d_{\text{out}} \geq 256$, the EMA write-back to persistent
global memory (required to retain updated statistics across optimizer steps) adds
an additional global-memory store that is not present in Stage 1 and exceeds the
savings from eliminating two separate EMA kernel launches. The per-element EMA
write-back cost grows linearly with layer width, and on the RTX~3060's wider
memory bus (336~GB/s) the effect is observable but small; the Fully Fused
Optimizer Step (Stage~3) resolves this by amortizing all memory traffic within a
single kernel invocation.


\begin{figure*}[t]
  \centering
  \includegraphics[width=\textwidth]{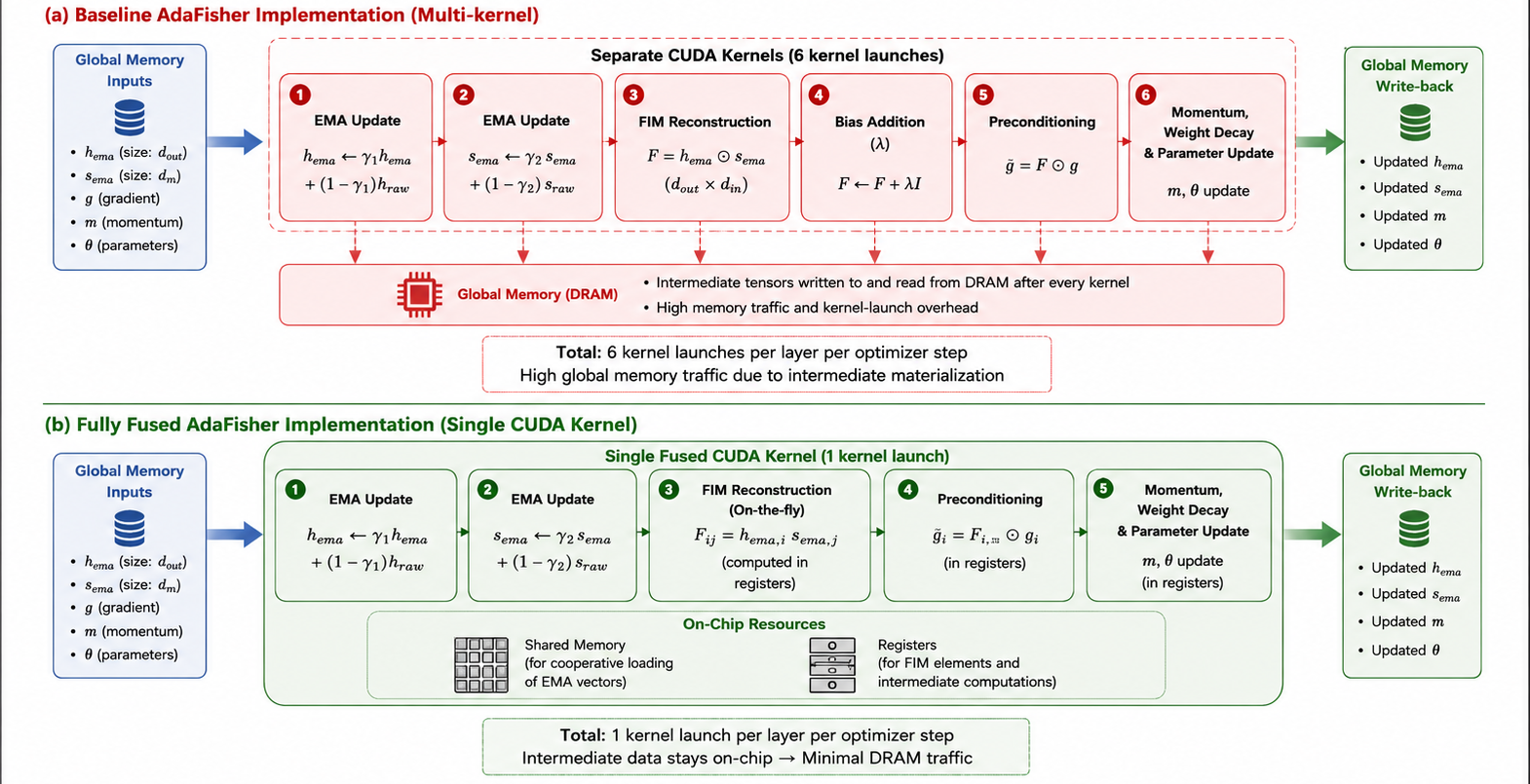}
  \caption{Comparison of execution flow between baseline and fully fused AdaFisher optimizers. The baseline uses multiple CUDA kernels with intermediate global memory transfers, while the fully fused implementation combines all operations into a single kernel, leveraging registers and shared memory to minimize DRAM traffic and kernel-launch overhead.}
  \label{fig:alg_compare}
\end{figure*}

\subsection{End-to-End Fused Optimizer Pipeline}
\label{subsec:fullyfused}

The third and final stage merges the entire optimizer pipeline---EMA updates, FIM
reconstruction, weight decay, momentum accumulation, and parameter update---into a
single CUDA kernel per layer. Algorithm~\ref{alg:full_fused} summarizes the fully
fused optimizer. The key enabling technique is loading the EMA vectors
$\hema$ and $\sema$ cooperatively into shared memory (lines 3--4), so that all
threads in a block reuse them without repeated DRAM accesses. The preconditioner
value $f_{ij}$ is computed in registers and consumed immediately; it is never
written to global memory. This reduces per-layer kernel launches from 6 (baseline)
to 1, and the visual comparison with the baseline is shown in
Figure~\ref{fig:alg_compare}.

For clarity, the fused kernel uses a consistent two-dimensional indexing convention. The index $i$ denotes the output-row dimension, with $0 \leq i < d_{\mathrm{out}}$, whereas $j$ denotes the input-column dimension, with $0 \leq j < d_{\mathrm{in}}$. Accordingly, $\hema[i]$ is associated with the output dimension and $\sema[j]$ with the input dimension. In the CUDA implementation, the thread coordinates are mapped as
\[
i = \mathtt{blockIdx.y} \times \mathtt{blockDim.y} + \mathtt{threadIdx.y},
\]
and
\[
j = \mathtt{blockIdx.x} \times \mathtt{blockDim.x} + \mathtt{threadIdx.x}.
\]
This mapping ensures consistency with the element-wise preconditioner definition
\[
\hat{F}_{ij} = \hema[i]\sema[j] + \lambda,
\]
particularly for layers where $d_{\mathrm{out}} \neq d_{\mathrm{in}}$.

\begin{algorithm}[t]
\caption{ Fully Fused Optimizer Step}
\label{alg:full_fused}
\begin{algorithmic}[1]

\Require $\hema, \sema, \hnew, \snew,
\nabla\theta, \theta, m,
\gamma_1, \gamma_2, \beta,
\eta, \lambda, \omega, b_c,
d_{\mathrm{out}}, d_{\mathrm{in}}$

\Ensure Updated $\theta$, $m$, $\hema$, and $\sema$

\State $\hema^{+}
\gets \gamma_1\hema + \gamma_2\hnew$

\State $\sema^{+}
\gets \gamma_1\sema + \gamma_2\snew$

\State Cooperatively load the required tiles of
$\hema^{+}$ and $\sema^{+}$ into shared memory

\State \Call{SyncThreads}{}

\ForAll{$(i,j)$ assigned to the current thread block}

    \State $f_{ij}
    \gets \hema^{+}[i]\sema^{+}[j]+\lambda$

    \State $g_{ij}
    \gets \nabla\theta_{ij}+\omega\theta_{ij}$

    \State $m_{ij}
    \gets \beta m_{ij}+(1-\beta)g_{ij}$

    \State $\theta_{ij}
    \gets \theta_{ij}
    -\dfrac{\eta}{b_c}
    \dfrac{m_{ij}}{f_{ij}}$

\EndFor

\State $\hema \gets \hema^{+}$
\State $\sema \gets \sema^{+}$

\end{algorithmic}
\end{algorithm}

Algorithm~\ref{alg:full_fused} summarizes the final CAFIN
stage, in which EMA updates, preconditioner construction,
weight decay, momentum accumulation, and parameter updates
are integrated into the fused optimizer execution path.

\subsection{Kernel Launch Configuration}
\label{subsec:kernel_config}

Table~\ref{tab:kernel_config} summarises the launch configuration and occupancy
for each custom CUDA kernel on the GTX~1660~Ti. All kernels use a 2-D thread
block of $16\!\times\!16 = 256$ threads, with the grid sized to cover the full
$d_{\text{out}} \times d_{\text{in}}$ output matrix. For the Fully Fused
Optimizer Step, each thread block loads a tile of the $\hema$ vector into
shared memory (16 floats, 64 bytes per tile) and a tile of $\sema$ (16 floats,
64 bytes per tile), for a total shared-memory usage of 576 bytes per block
including synchronization overhead. Theoretical occupancy is computed from the
CUDA occupancy calculator using the register counts reported by
\texttt{ptxas}; no register spilling is observed for any kernel at these block
sizes on sm\_75.

\begin{table}[t]
\centering
\caption{Kernel Launch Configuration and Occupancy (GTX~1660~Ti, sm\_75)}
\label{tab:kernel_config}
\footnotesize
\setlength{\tabcolsep}{3pt}
\begin{tabular}{@{}lcccc@{}}
\toprule
\textbf{Kernel} & \textbf{Block} & \textbf{Regs/thread} & \textbf{Smem (B)} & \textbf{Occ.} \\
\midrule
Eager baseline (\texttt{kron})        & N/A     & 12 & 0   & 100\% \\
Fused Preconditioner                   & $16\!\times\!16$ & 20 & 576 & 100\% \\
Fused EMA + Prec.                      & $16\!\times\!16$ & 24 & 576 & 100\% \\
Fully Fused Optimizer Step             & $16\!\times\!16$ & 28 & 576 & 100\% \\
\bottomrule
\end{tabular}
\end{table}

\subsection{Complexity Analysis}
\label{subsec:complexity}

The asymptotic arithmetic complexity remains $O(d_{\text{out}}\cdot d_{\text{in}})$.
The fusion strategy reduces per-layer kernel launches from 6 to 1, eliminates
intermediate tensor materialization, reduces DRAM traffic, and leverages shared
memory and register-resident computation, improving execution efficiency without
altering AdaFisher's mathematical formulation.

\subsection{Numerical Correctness}
\label{subsec:numerics}

To validate numerical equivalence between the eager and fused implementations, we compute the relative Frobenius norm of parameter updates $\Delta W$ across all layers:

\begin{equation}
\label{eq:frobenius_error}
\text{Error} =
\frac{
\left\|
\Delta W_{\text{baseline}}
-
\Delta W_{\text{fused}}
\right\|_{F}
}{
\left\|
\Delta W_{\text{baseline}}
\right\|_{F}
}
\end{equation}

Using the relative error defined in
Equation~\eqref{eq:frobenius_error}, we evaluate numerical
consistency across all Conv2d and Linear layers.
Across CIFAR-10 and CIFAR-100, all evaluated layers remain
within the predefined numerical tolerance of
$2\times10^{-6}$, indicating numerical consistency between
the baseline and fused FP32 implementations.

For CIFAR-10, the maximum observed deviation is 
$1.663 \times 10^{-6}$, while CIFAR-100 reaches 
$1.858 \times 10^{-6}$, both within the defined threshold. 

This confirms that kernel fusion preserves optimizer semantics without introducing numerical drift.

\begin{figure}[h]
\centering
\includegraphics[width=1\linewidth]{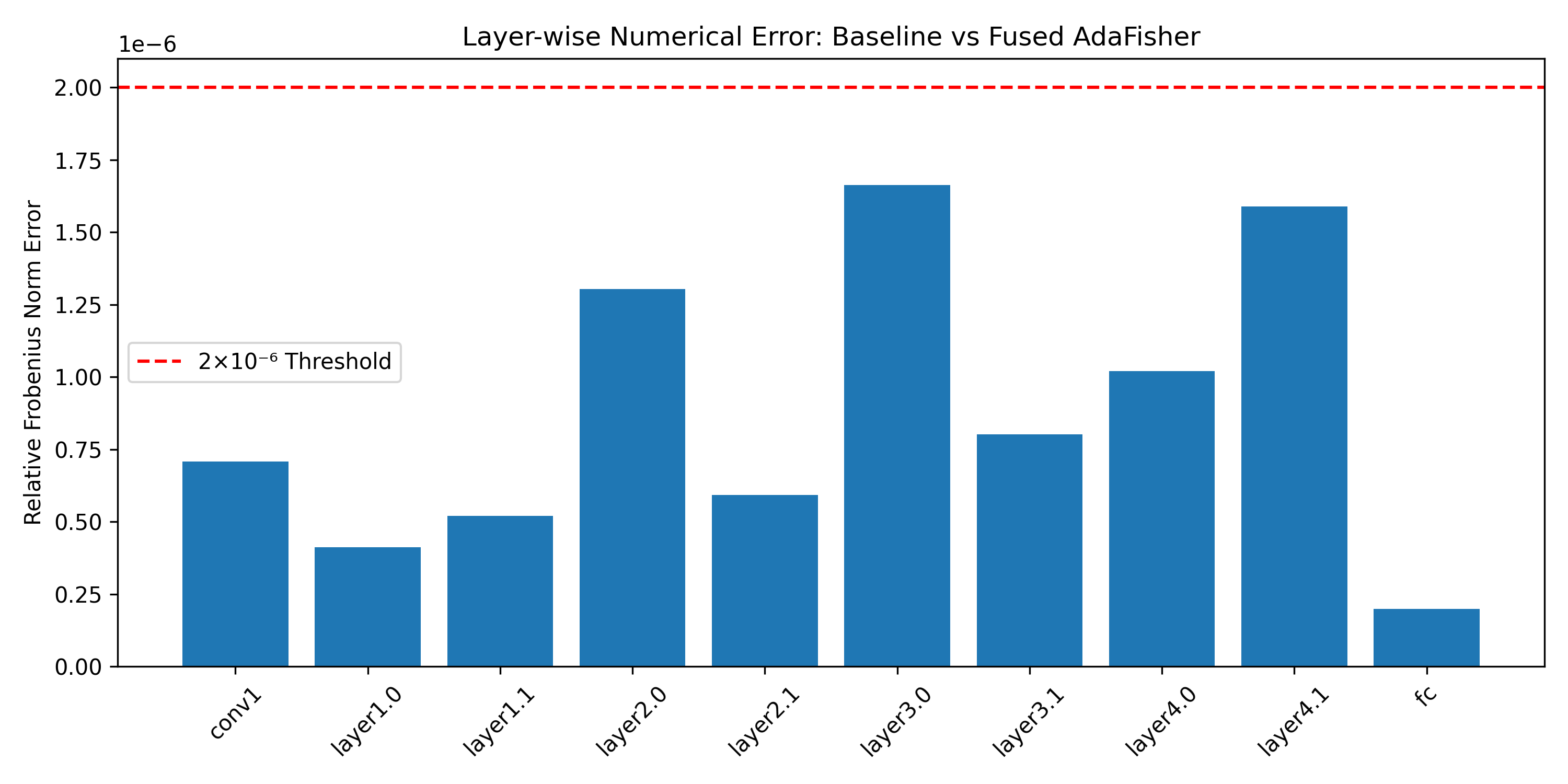}
\caption{Layer-wise Frobenius norm error between baseline and fully fused AdaFisher implementation on CIFAR-10. The dashed line represents the unified numerical tolerance threshold ($2 \times 10^{-6}$). All layers remain within the predefined numerical tolerance, confirming numerical consistency between the baseline and fully fused FP32 implementations.}
\label{fig:frobenius_error}
\end{figure}

The layer-wise numerical deviations are visualized in
Figure~\ref{fig:frobenius_error}, while
Table~\ref{tab:numerical_correctness} summarizes the maximum
Frobenius-norm errors observed across CIFAR-10 and CIFAR-100.

\begin{table}[h]
\centering
\caption{Numerical correctness across datasets under the predefined FP32 numerical tolerance ($2\times10^{-6}$).}
\label{tab:numerical_correctness}
\begin{tabular}{lccc}
\hline
Dataset & Max Frobenius Error & Threshold & Status \\
\hline
CIFAR-10 & $1.663 \times 10^{-6}$ & $2 \times 10^{-6}$ & PASS \\
CIFAR-100 & $1.858 \times 10^{-6}$ & $2 \times 10^{-6}$ & PASS \\
\hline
\end{tabular}
\end{table}
\subsection{PyTorch Integration}
\label{subsec:integration}

CUDA kernels are compiled using PyTorch \texttt{CppExtension} with
\texttt{nvcc -O3 --use\_fast\_math}. Python-callable entry points are exposed
through \texttt{pybind11}. The surrounding PyTorch training loop remains unchanged.

\section{Experimental Evaluation}
\label{sec:experimental_evaluation}

This section explains how the experiments were set up. It covers the
questions we wanted to answer, the hardware and software used, the
datasets and model, the hyperparameters, the execution modes we compared,
and how we kept the experiments reproducible. Results and analysis are
presented separately, in Section~\ref{sec:statistical_performance}.

\subsection{Research Questions}
\label{sec:research_questions}

The experimental evaluation is designed to answer the following questions:

\begin{itemize}
  \item \textbf{RQ1:} How much does each fusion stage reduce isolated Fisher
    Information Matrix (FIM) construction latency and end-to-end optimizer-step
    latency on consumer-grade GPUs?
  \item \textbf{RQ2:} Does kernel fusion preserve convergence accuracy across
    multiple datasets and random seeds?
  \item \textbf{RQ3:} How much peak GPU memory is reduced through intermediate
    tensor elimination?
\end{itemize}

\subsection{Hardware and Software Environment}
\label{sec:hardware_software}

Two GPU architectures are evaluated to determine whether fusion benefits remain
consistent across Turing and Ampere microarchitectures, which differ in cache
capacity, shared-memory bandwidth, and occupancy behaviour. Hardware specifications
are summarised in Table~\ref{tab:hw}.

\begin{table}[t]
\centering
\caption{Hardware Configuration}
\label{tab:hw}
\begin{tabular}{@{}lcccc@{}}
\toprule
\textbf{GPU} & \textbf{Arch.} & \textbf{Cores}
             & \textbf{VRAM} & \textbf{BW} \\
\midrule
GTX 1660~Ti & Turing (sm\_75)  & 1536 & 6.4\,GB & 288\,GB/s \\
RTX 3060 L  & Ampere (sm\_86)  & 3840 & 6.4\,GB & 336\,GB/s \\
\bottomrule
\end{tabular}
\end{table}

Unless otherwise stated, isolated FIM latency measurements, optimizer-step latency
experiments, and CIFAR convergence evaluations are conducted on the GTX~1660~Ti.
Tiny ImageNet cross-mode comparisons Table~\ref{tab:tiny_imagenet} are evaluated on
the RTX~3060 Laptop GPU for 50 epochs under identical hardware conditions across all
execution modes. The 200-epoch Tiny ImageNet result for the Fully Fused Optimizer
Step on the GTX~1660~Ti is reported separately in Table~\ref{tab:ffs}.
Cross-architecture multi-seed optimizer-step latency results are also reported in
Table~\ref{tab:ffs}.

Experiments are implemented using PyTorch~2.6.0+cu124, Python~3.10,
CUDA~12.4/12.6, and Windows~10/11. CUDA kernels are compiled with
\texttt{nvcc -O3 --use\_fast\_math}, using architecture flags
\texttt{-arch=sm\_75} for the GTX~1660~Ti and \texttt{-arch=sm\_86} for the
RTX~3060 Laptop GPU. The host compiler is MSVC~14.36 via
\texttt{torch.utils.cpp\_extension}.

\subsection{Datasets and Model}
\label{sec:datasets_model}

Experiments are conducted on CIFAR-10 and
CIFAR-100~\cite{krizhevsky2009} ($32\!\times\!32$) and Tiny
ImageNet~\cite{le2015} ($64\!\times\!64$, 200 classes). Standard data augmentation
is applied using random crop with 4-pixel padding and random horizontal flip.
Cutout regularisation uses one hole of length 16 pixels. Dataset-specific
per-channel mean and standard deviation normalisation is applied to all datasets.

All experiments use ResNet-18~\cite{he2016resnet} with a CIFAR-adapted first
convolution layer ($3\!\times\!3$, stride-1, max-pooling removed). For Tiny
ImageNet, the initial stride is increased to accommodate the larger
$64\!\times\!64$ input resolution.

\subsection{Hyperparameters}
\label{sec:hyperparameters}

All execution modes are trained with the same settings. This keeps the
comparison fair, so any speed difference comes from the fusion strategy
and not from a different training setup. Table~\ref{tab:hyp} lists all
hyperparameter values used across experiments.

\begin{table}[t]
\centering
\caption{Hyperparameter Configuration}
\label{tab:hyp}
\begin{tabular}{@{}lc@{}}
\toprule
\textbf{Hyperparameter} & \textbf{Value} \\
\midrule
Learning rate $\eta$    & 0.001 \\
Momentum $\beta$        & 0.9 \\
EMA decay $\gamma$      & 0.8 \\
Damping $\lambda$       & 0.001 \\
Weight decay $\omega$   & $5\!\times\!10^{-4}$ \\
Batch size              & 256 \\
Epochs                  & 200 (CIFAR) / 50 (Tiny ImageNet) \\
LR schedule             & CosineAnnealingLR \\
TCov interval           & 100 steps \\
Seeds                   & 42, 123, 456 \\
Precision               & FP32 \\
\bottomrule
\end{tabular}
\end{table}

TCov denotes the Fisher statistics update interval. EMA statistics are recomputed
every 100 training steps and remain fixed between updates.

\subsection{Execution Modes}
\label{sec:execution_modes}

Table~\ref{tab:modes} summarises the four execution modes evaluated in this work.
Each mode corresponds to a successive fusion stage, as illustrated in
Figure~\ref{fig:pipeline}. All modes use identical optimizer states and
hyperparameters, isolating the effect of execution-level fusion from algorithmic
variation.

\begin{table}[t]
\centering
\caption{Execution Modes Compared}
\label{tab:modes}
\begin{tabular}{@{}p{2.8cm}p{5.2cm}@{}}
\toprule
\textbf{Mode} & \textbf{Description} \\
\midrule
Eager Baseline
  & Baseline PyTorch implementation using \texttt{kron()} and separate tensor
    operations \\[3pt]
Fused Preconditioner
  & Single fused outer-product CUDA kernel for FIM construction \\[3pt]
Fused EMA + Prec.
  & EMA update fused with outer-product computation \\[3pt]
Fully Fused Optimizer
  & Entire optimizer step fused into a single CUDA kernel \\
\bottomrule
\end{tabular}
\end{table}

\subsection{Reproducibility}
\label{sec:reproducibility}

All experiments use fixed random seeds for data loading, model initialisation, and
CUDA operations. Deterministic cuDNN execution is enabled via
\texttt{torch.backends.cudnn.deterministic=True}, while benchmarking mode is
disabled to prevent algorithm-selection variability across runs. All reported
timing values reflect steady-state performance measured after a dedicated GPU
warmup pass. The first training epoch is excluded from all timing measurements to
remove CUDA JIT compilation effects.

\section{Statistical and Performance Analysis}
\label{sec:statistical_performance}

This section presents what we measured and what it means. We first explain
each metric. Then we check that our baseline is a correct reproduction of
AdaFisher. After that, we look at kernel-level, memory, and end-to-end
results, followed by the statistical tests, the practical cost savings,
and finally the numerical and convergence checks.

\subsection{Performance and Evaluation Metrics}
\label{sec:evaluation_metrics}

These metrics cover a range of aspects: FIM construction latency, optimizer-step latency, distributed latency, CPU usage, and memory consumption.
Memory required of the GPU, total training time and the accuracy (numerical error).
They are used to answer various questions.

\textbf{Convergence.}
The evaluation is done by the accuracy of the Top-1 test on the held out evaluation set, and convergence is measured by this.
The last train epoch. Figure~\ref{fig:convergence} reports accuracy
eager preliminary and fused preconditioner with three random trajectories.No. 3 for the 2 different random scenarios, Eager Baseline and Fused Preconditioner.
Next, your answer is validated with kernel fusion on CIFAR-100, thus making sure the dynamics of the learning remain the same.

\textbf{FIM construction latency.}
Isolates per-layer preconditioner construction time (synchronized)
Timing (all ResNet-18 layers) of CUDA events (during training). The first training
To avoid CUDA JIT compilation overhead, epoch is not considered in timing.
This is a measure of the acceleration of the step itself by the FIM-fusion step.

\textbf{Optimizer-step latency.}
The wall-clock time for optimizer.step() is timed using
\texttt{torch.cuda.synchronize()} barriers. Data presented as
You could use the mean of the per-step measurements where the value is
Is less than $3*mdn$. This threshold is a single filter that provides isolation conditioning.
timings, making virtual machines appear faster.overestimate of individual timings due to thermal-throttling hiccup and OS schedular disruption.
This factor, 3$\times$, represents measurements taken, not in the steady state of the GPU, but rather when the device is first powered on.
was selected empirically as one that removes less than 2
Measures for capturing the true extreme value points. This indicates the true end-to-end advantage for this metric.
Not only the single FIM benefit, but also a fusion benefit.

\textbf{Peak GPU memory.}
All memory used by GPU is tracked during the training process with
\texttt{torch.cuda.max\_memory\_allocated()}. Peak tracking is reset once the
Pass warmup to remove the initialisation burden of the once-only set up.

Numerical error and accuracy.
To ensure that doing the fusion does not alter what AdaFisher actually computes, we
Compare parameter updates and terminal test accuracy of baseline and baseline plus;Compare the parameter updates and the end test accuracy for baseline and baseline plus;
A new approach, called simply fused implementations, is described in Sections~\ref{sec:statistical_significance}
and~\ref{sec:convergence_equivalence}.

\subsection{Reproduction Validation}
\label{sec:reproduction_validation}

We always first verify that our own "baseline" is correct, before comparing fusion stages.
AdaFisher implementation actually acts the same as the original AdaFisher
paper reports. This step is important since it is a speed comparison that is only done once
meaningful only if it's a correct reference implementation.

To compare the accuracy in our baseline model with the published AdaFisher results, we compare it to model numbers in Table~\ref{tab:repro}. The increase in accuracy is due to the Cutout augmentation and the first convolution of the $3\times3$ stride-1 that were adapted, and the results are 2.37 ppts above(=) the published value in our CIFAR-10 accuracy. The CIFAR-100 test result is within one standard deviation of the test result. In these results, we were able to prove that our baseline was a faithful and correct reproduction of the original optimizer.

Contributions to the Copernicus emergency response service are also provided using this Travel Gateway, using which we run a Two
A novel method of One-Sided Test (TOST) for equivalence analysis is used on the CIFAR-10 and CIFAR-100 data sets.A novel One-Sided Test (TOST) is applied to the CIFAR-10 and CIFAR-100 data sets.
accuracies. TOST's pressing question is: Are the two sets of results
Close enough relative to be considered equivalent within a pre-set, small margin?
The margin is anticipated to be in the range of $\pm 0.1\%$ as per the conventionally accepted practice in deep oils.
learning reproducibility studies. A detailed description of this test,
and how it differs from a plain significance test, is given in
Section~\ref{sec:statistical_significance}.

\begin{figure}[t]
\centering
\includegraphics[width=1.1\linewidth]{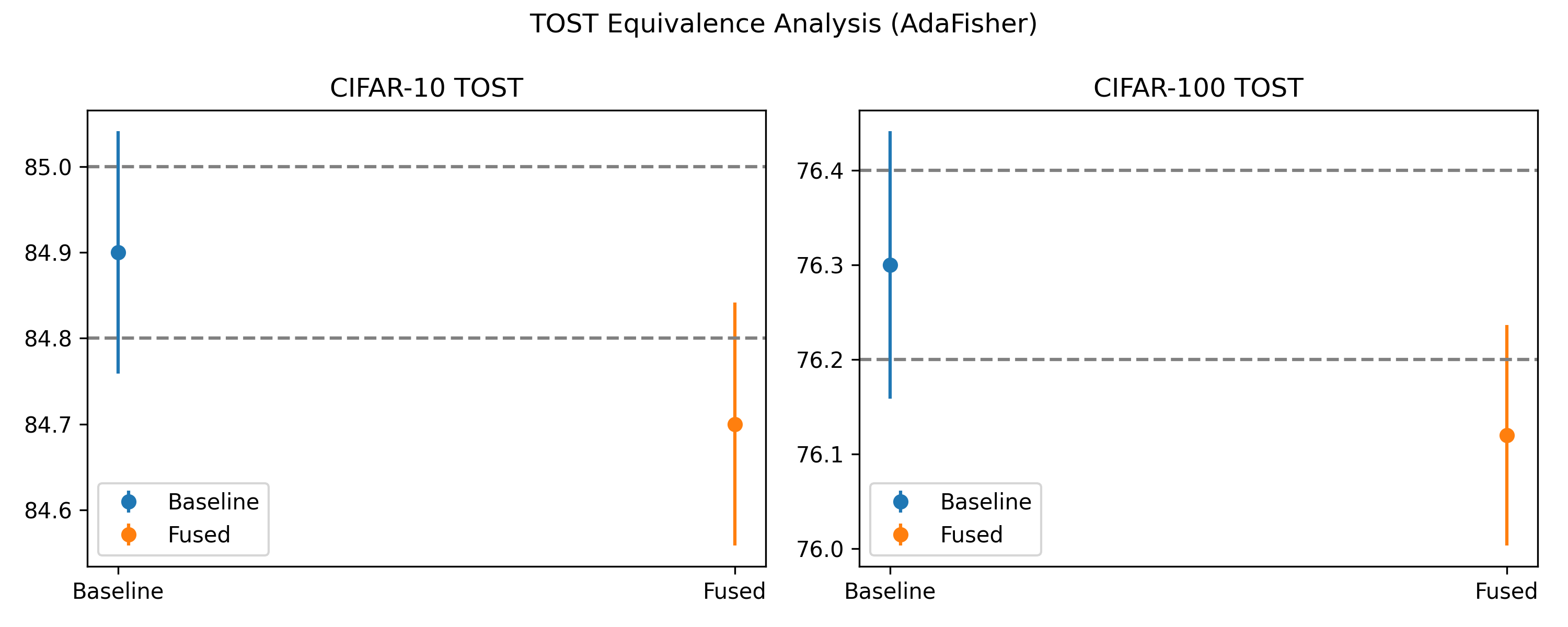}
\caption{TOST equivalence analysis across CIFAR-10 and CIFAR-100. The shaded region represents the equivalence margin ($\pm 0.1\%$). Overlapping confidence intervals indicate statistical equivalence of baseline and fused optimization behavior across multiple seeds.}
\label{fig:tost}
\end{figure}

The results of the Tost equivalence are displayed in the figure~\ref{fig:tost}.
for three random seed sets (for CIFAR-10 and CIFAR-100),
illustrating how the observed accuracy is related to the...
Differences and the preformed equivalence margin.

\begin{table}[t]
\centering
\caption{Reproduction Validation: Baseline Accuracy (\%)}
\label{tab:repro}
\begin{tabular}{@{}lcccc@{}}
\toprule
\textbf{Dataset} & \textbf{Paper}
  & \textbf{Seed 42} & \textbf{Seed 123} & \textbf{Seed 456} \\
\midrule
CIFAR-10  & 93.97 & 96.34 & 96.27 & 96.35 \\
CIFAR-100 & 77.17 & 77.39 & 77.24 & 77.46 \\
\bottomrule
\end{tabular}
\end{table}

\subsection{Kernel-Level Hardware Analysis}
\label{subsec:exp1}

This part looks at why the GPU behaves the way it does, using compiler
output and a roofline model.

From \texttt{ptxas} compiler output on the GTX~1660~Ti, the Fused Preconditioner
kernel uses 20 registers per thread to keep intermediate values on-chip, compared
to 12 registers for the naive baseline and 10 for the algebraically simplified
variant. Despite the higher register count, it maintains 100\% theoretical occupancy
by using 576~bytes of shared memory per block, while the naive baseline requires
0~bytes of shared memory and the algebraic variant uses 192~bytes with one
synchronisation barrier. Full launch parameters and occupancy figures are
summarised in Table~\ref{tab:kernel_config}.

\begin{figure}[t]
  \centering
  \includegraphics[width=\columnwidth]{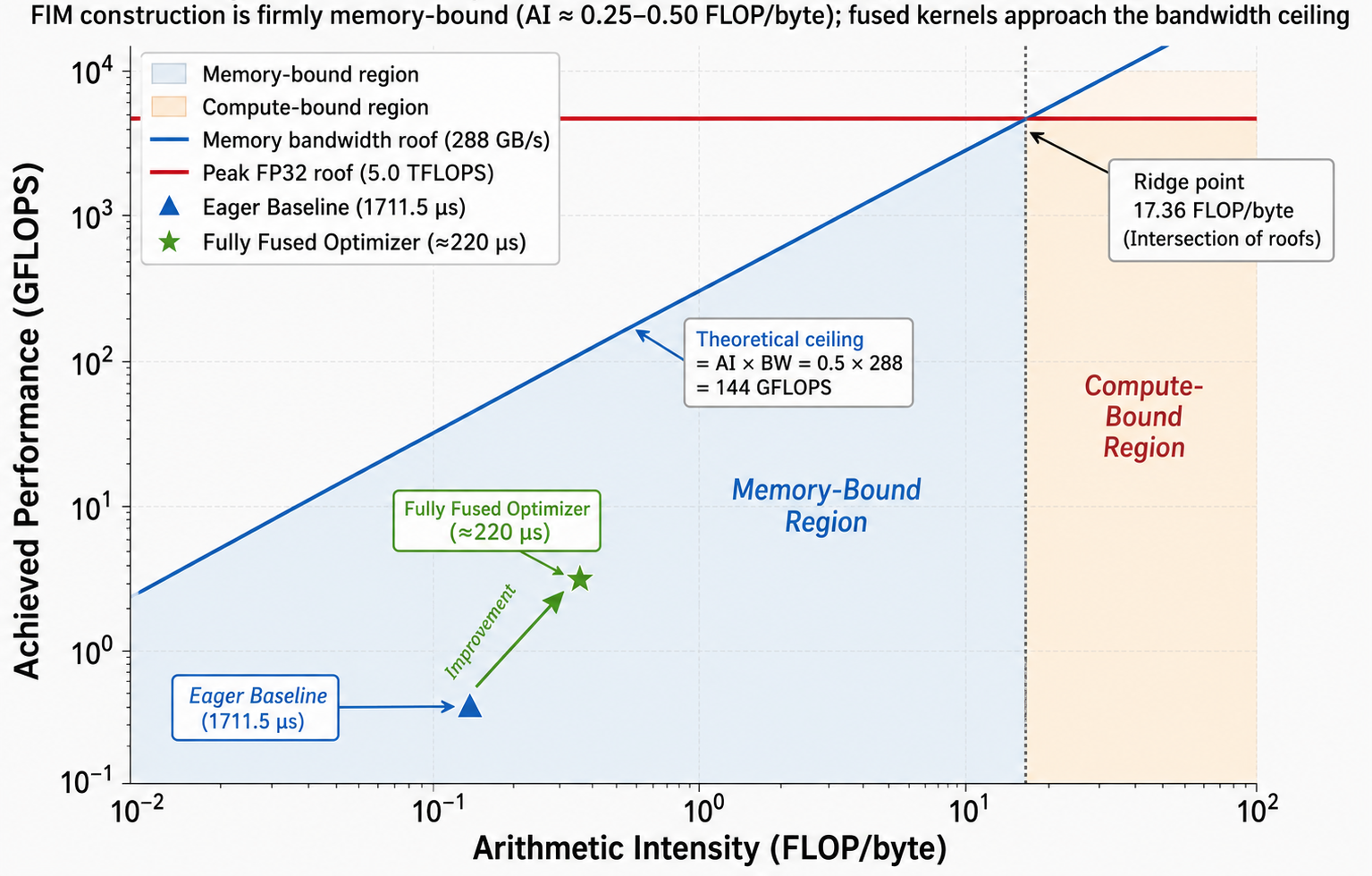}
  \caption{Roofline analysis showing that the fully fused AdaFisher reduces DRAM transfers, increases arithmetic intensity, and achieves higher GPU performance compared to the baseline.}
  \label{fig:roofline}
\end{figure}

Figure~\ref{fig:roofline} plots all four execution modes on the GTX~1660~Ti
roofline. The peak bandwidth is 288.0~GB/s and peak FP32 throughput is 5.0~TFLOPS,
giving a ridge point of 17.36~FLOP/byte. The FIM kernel has an arithmetic intensity
of $\approx\!0.5$~FLOP/byte, which lies far below the ridge, placing it firmly in
the memory-bound regime. The theoretical performance ceiling is
$0.5\!\times\!288.0 = 144$~GFLOPS. The Fused Preconditioner achieves
$\approx\!1.59$~GFLOPS on layer4.1, compared to 0.31~GFLOPS for the baseline,
confirming that fusion moves execution closer to the bandwidth ceiling.

\subsection{Isolated FIM Construction Performance}
\label{subsec:exp2}

This experiment isolates just the FIM-construction step, so we can see
the direct effect of fusing that one operation, separate from the rest
of the optimizer.

\begin{figure*}[t]
  \centering
  \includegraphics[width=\textwidth]{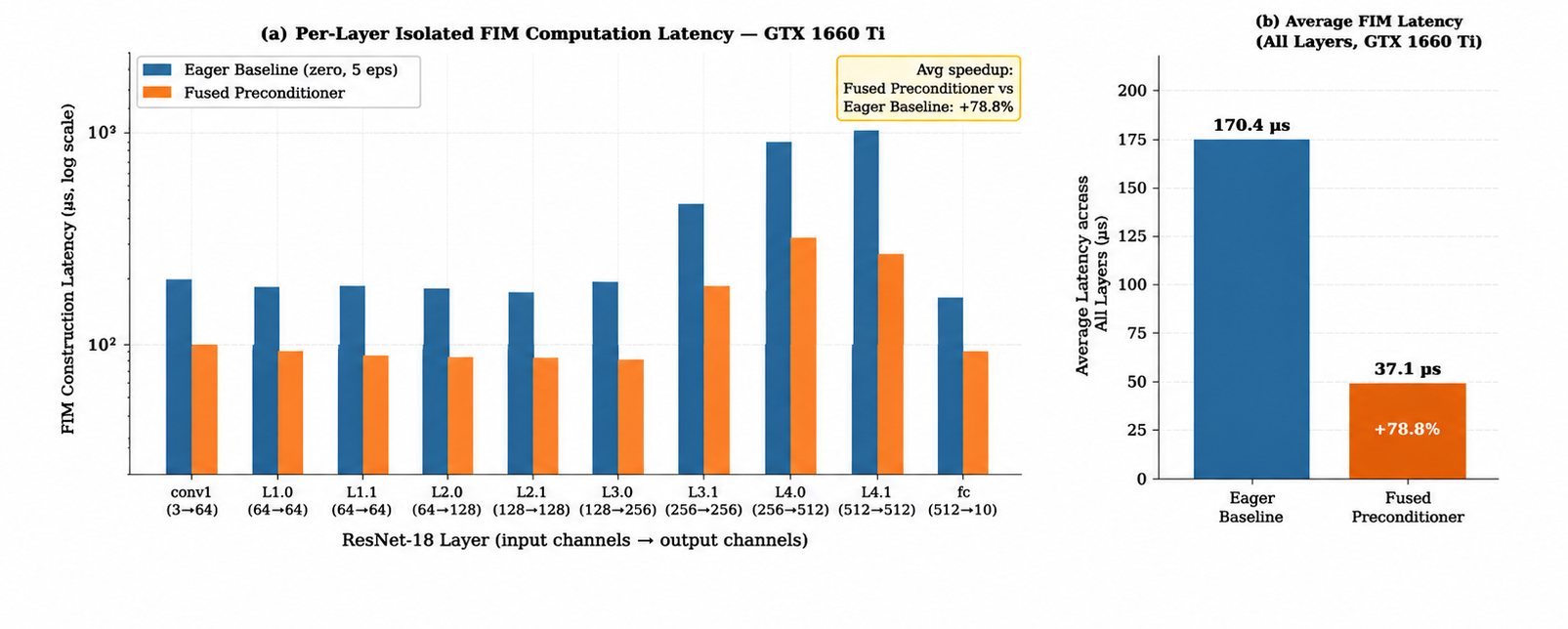}
  \caption{Per-layer FIM construction latency (GTX~1660~Ti, log scale).
    \textit{(a)}~Latency across all ten layers; the Fused Preconditioner reduces
    average latency from 170.4\,\textmu s to 37.1\,\textmu s.
    \textit{(b)}~Mean latency showing progressive speedup at each fusion stage.}
  \label{fig:perlayer}
\end{figure*}

Figure~\ref{fig:perlayer} reports per-layer FIM construction latency on the
GTX~1660~Ti. The Fused Preconditioner achieves an average speedup of \textbf{78.4\%}
over the eager baseline and \textbf{60.8\%} over the algebraic simplification across
all ten ResNet-18 layers. Table~\ref{tab:exp2} in Appendix~\ref{app:perlayer} provides the complete numerical breakdown per layer.

\subsection{GPU Memory Savings}
\label{sec:memory_savings}

Fusion also cuts down GPU memory use, because it avoids writing the
full FIM matrix to memory in the first place.

\begin{figure*}[t]
  \centering
  \includegraphics[width=\textwidth]{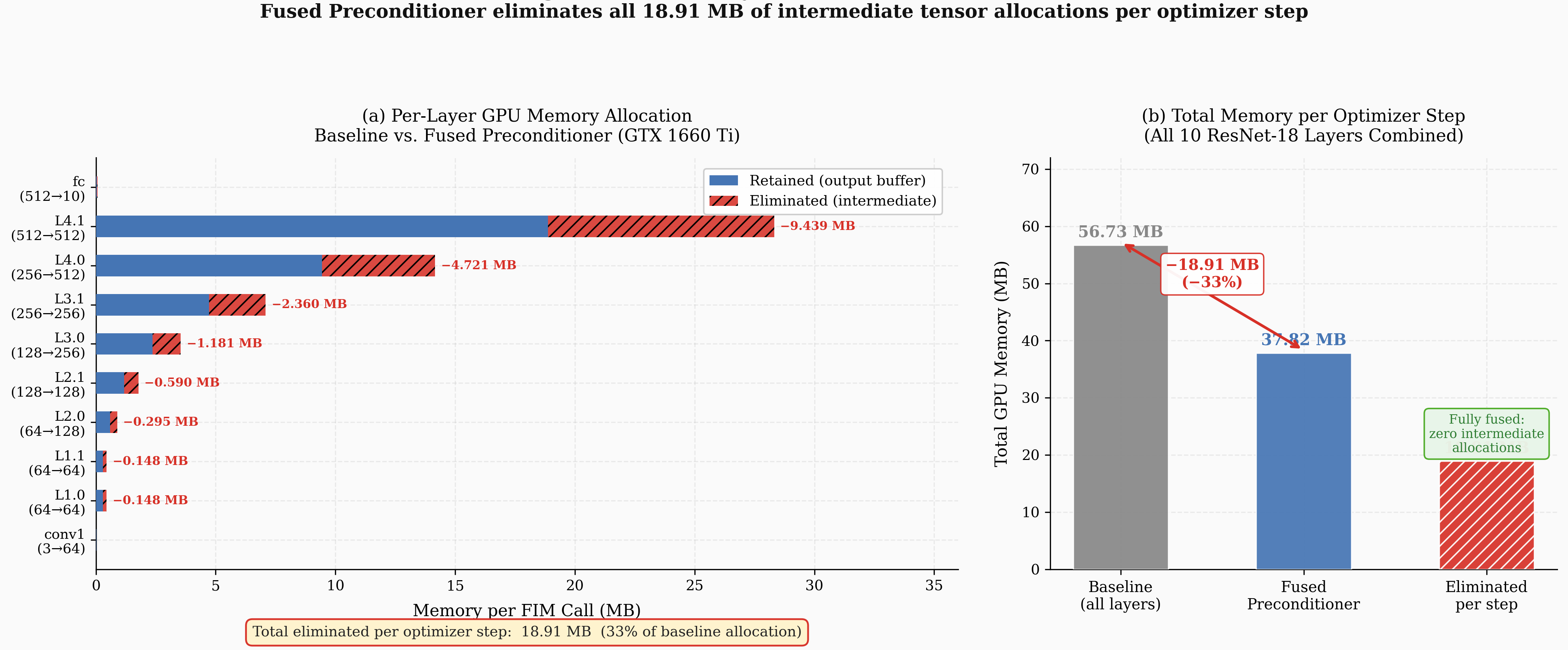}
  \caption{GPU memory usage per FIM call for baseline and fused AdaFisher. The fused preconditioner eliminates intermediate tensor allocations, reducing total memory per optimizer step by 18.91 MB.}
  \label{fig:memory}
\end{figure*}

The Fused Preconditioner eliminates all intermediate FIM tensor allocations. The
total saving is \textbf{18.91~MB} per optimizer step across all ten layers. For
layer4.1 alone, 9.44~MB is saved per call. Figure~\ref{fig:memory} shows the
per-layer breakdown. Table~\ref{tab:memory_stages} summarises peak GPU memory
usage across all four execution modes during 200-epoch training on CIFAR-100
(GTX~1660~Ti), including model parameters, optimizer state, activations, and
intermediate allocations.

\begin{table}[t]
\centering
\caption{Peak GPU Memory Across Execution Modes (CIFAR-100, GTX~1660~Ti)}
\label{tab:memory_stages}
\footnotesize
\setlength{\tabcolsep}{4pt}
\begin{tabular}{@{}lcc@{}}
\toprule
\textbf{Mode} & \textbf{Peak Mem (MB)} & \textbf{vs Baseline} \\
\midrule
Eager Baseline              & 1247 & --- \\
Fused Preconditioner        & 1228 & $-$19\,MB \\
Fused EMA + Prec.           & 1228 & $-$19\,MB \\
Fully Fused Optimizer Step  & 1228 & $-$19\,MB \\
\bottomrule
\end{tabular}
\end{table}

\subsection{End-to-End Optimizer Performance}
\label{sec:end_to_end}

The isolated FIM result shows the benefit of one piece of the optimizer.
Here we check whether that benefit shows up in the complete optimizer
step, and across different GPUs, datasets, and seeds.

\begin{figure*}[t]
  \centering
  \includegraphics[width=\textwidth]{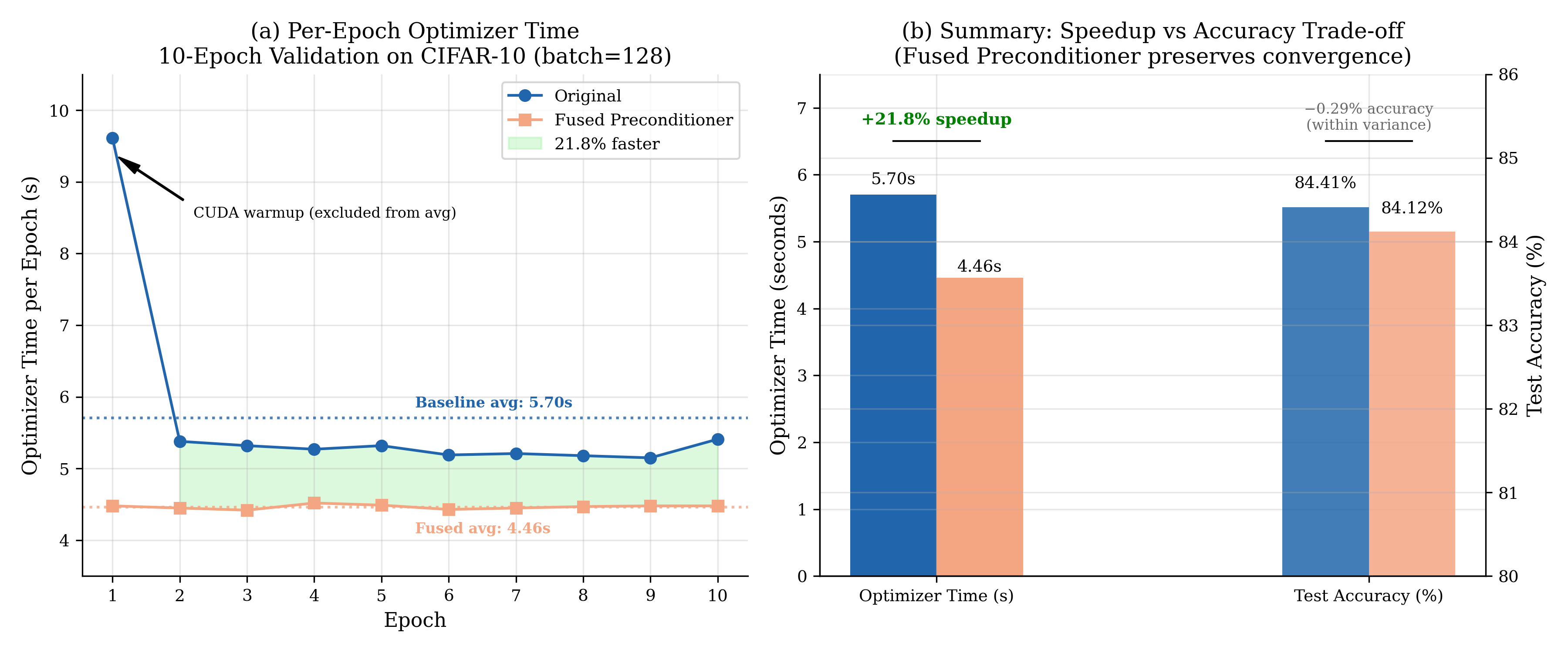}
  \caption{10-epoch validation on CIFAR-10 (batch=128). \textit{Left}: per-epoch optimizer time. \textit{Right}: summary speedup vs.\ accuracy trade-off showing 21.8\% speedup while preserving convergence.}
  \label{fig:10epoch}
\end{figure*}

\begin{figure}[t]
  \centering
  \includegraphics[width=\columnwidth]{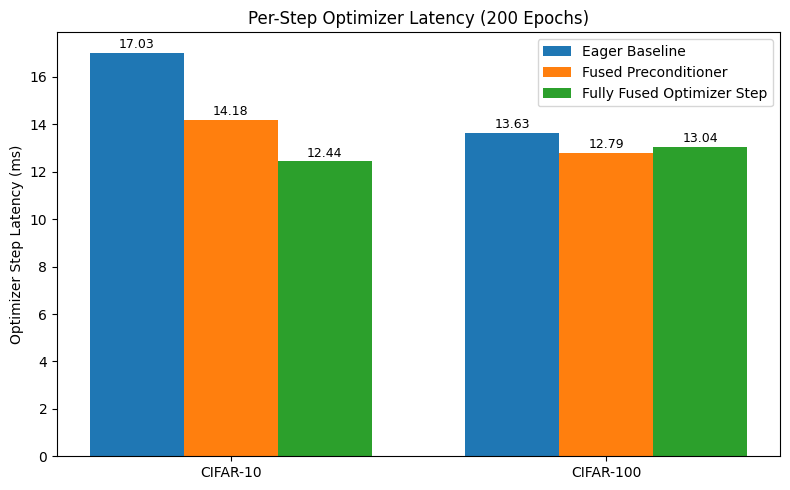}
  \caption{Optimizer step latency for baseline, fused preconditioner, and fully fused AdaFisher on GTX 1660 Ti. Fusion reduces per-step latency, with the fully fused optimizer achieving an additional 1.5–4.3 percent speedup.}
  \label{fig:latency}
\end{figure}

Table~\ref{tab:exp3} reports the 10-epoch CIFAR-10 validation
results on the GTX~1660~Ti with a batch size of 128. The Fused
Preconditioner achieves a 21.8\% reduction in optimizer-step time
with an accuracy difference of only $-0.29\%$ relative to the
baseline. Figure~\ref{fig:10epoch} visualizes the corresponding
per-epoch optimizer time and the speedup--accuracy trade-off.
Figure~\ref{fig:latency} further compares optimizer-step latency
across the evaluated fusion stages.

\begin{figure}[h]
\centering
\includegraphics[width=0.7\linewidth]{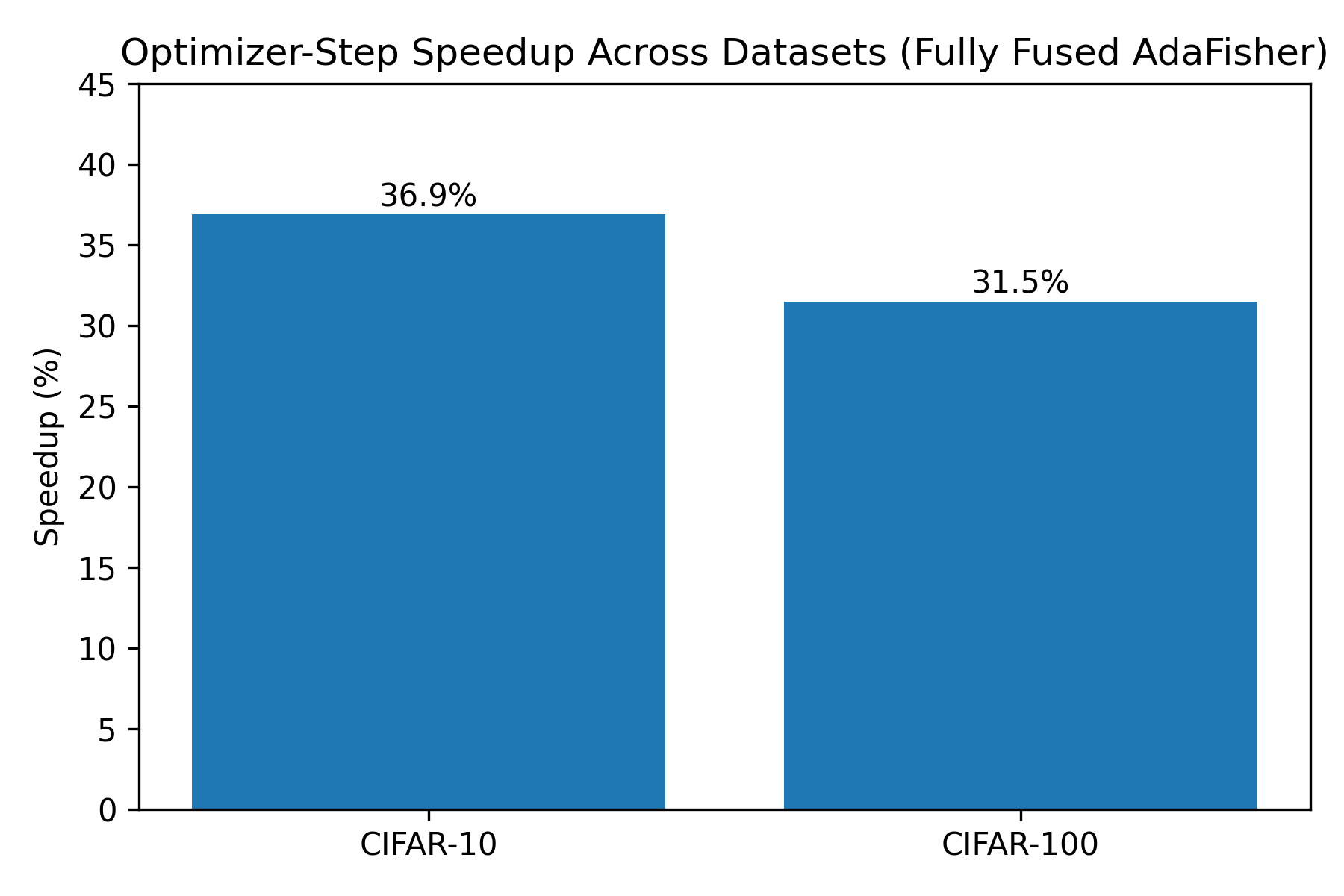}
\caption{Optimizer-step speedup across datasets for the fully fused AdaFisher kernel. The results show consistent latency reduction on CIFAR-10 and CIFAR-100, demonstrating robustness of kernel fusion across dataset scales.}
\label{fig:speedup_datasets}
\end{figure}

Table~\ref{tab:ffs} reports the 200-epoch results for the Fully
Fused Optimizer Step across both GPUs and multiple random seeds.
The kernel reduces per-step optimizer latency by 2.1--26.9\%
while maintaining comparable convergence accuracy.
Figure~\ref{fig:speedup_datasets} summarizes the optimizer-step
speedup across CIFAR-10 and CIFAR-100, illustrating the consistency
of the performance improvement across dataset scales.

Optimizer latency is similar across both GPUs ($\approx\!11$--17\,ms) despite a
factor of $\sim\!2$ difference in total step time, confirming that optimizer-step
execution is bandwidth-limited rather than compute-limited.

\begin{table}[t]
\centering
\caption{Fully Fused Optimizer Step: 200-Epoch Results (Opt(clean) in ms)}
\label{tab:ffs}
\footnotesize
\setlength{\tabcolsep}{3pt}
\begin{tabular}{@{}llccc@{}}
\toprule
\textbf{Dataset/GPU/Seed} & \textbf{Mode} & \textbf{Acc (\%)} & \textbf{Opt (ms)} & \textbf{Spd.} \\
\midrule
C-10/GTX1660Ti/s=42  & Eager Baseline   & 96.27 & 17.03 & ---       \\
                     & Fully Fused Opt. & 96.02 & 12.44 & $+26.9\%$ \\[2pt]

C-10/RTX3060/s=456   & Eager Baseline   & 96.35 & 11.58 & ---       \\
                     & Fully Fused Opt. & 96.10 & 11.34 & $+2.1\%$  \\[2pt]

C-100/GTX1660Ti/s=42 & Eager Baseline   & 77.19 & 13.63 & ---       \\
                     & Fully Fused Opt. & 76.61 & 13.04 & $+4.3\%$  \\[2pt]

C-100/RTX3060/s=123  & Eager Baseline   & 76.42 & 12.50 & ---       \\
                     & Fully Fused Opt. & 75.95 & 11.46 & $+8.3\%$  \\
\bottomrule
\end{tabular}
\end{table}

\begin{table}[h]
\centering
\caption{Tiny ImageNet evaluation across execution modes (RTX 3060, 50 epochs).}
\label{tab:tiny_imagenet}
\begin{tabular}{lccc}
\hline
Mode & Top-1 Accuracy (\%) & Step Time (ms) & Speedup \\
\hline
Eager Baseline & 61.42 & 18.90 & -- \\
Fused Preconditioner & 61.10 & 16.01 & +15.3\% \\
Fused EMA + Prec. & 60.85 & 19.55 & -3.5\% \\
Fully Fused Optimizer & 60.98 & 18.02 & +4.6\% \\
\hline
\end{tabular}
\end{table}

Table~\ref{tab:tiny_imagenet} compares all four modes on Tiny
ImageNet (RTX~3060, 50 epochs, seed=42). The Fused Preconditioner achieves the
strongest optimizer step reduction ($+15.3\%$). The Fused EMA + Preconditioner
shows a minor regression ($-3.5\%$) due to EMA write-back overhead on longer
layers, as discussed in Section~\ref{subsec:ema_integration}. The Fully Fused
Optimizer Step achieves $+4.6\%$. Accuracy differences across all modes are
within expected short-run variance.

\subsection{Statistical Significance}
\label{sec:statistical_significance}

This subsection checks whether the differences we measured are real, or
whether they could just be normal run-to-run noise. 

We use two statistical analyses corresponding to two different objectives.
Welch's t-test is used to determine whether the observed optimizer-step
latency reduction is statistically significant, while TOST equivalence
testing is used to verify that accuracy differences between baseline and
fused implementations remain within an acceptable equivalence margin.

The two tests we use answer two different
questions, and it helps to keep them separate:

\begin{itemize}
  \item \textbf{Welch's $t$-test} asks: is there a statistically
    detectable \emph{difference} between two sets of measurements
    (for example, baseline vs.\ fused optimizer-step latency)?
  \item \textbf{TOST} asks the opposite kind of question: are two sets
    of measurements close enough to call them \emph{equivalent}, within
    a small margin set in advance? A result can fail to show a
    significant difference under a $t$-test without actually proving
    equivalence; TOST is the more direct way to check equivalence.
\end{itemize}

\begin{figure}[t]
\centering
\includegraphics[width=0.85\linewidth]{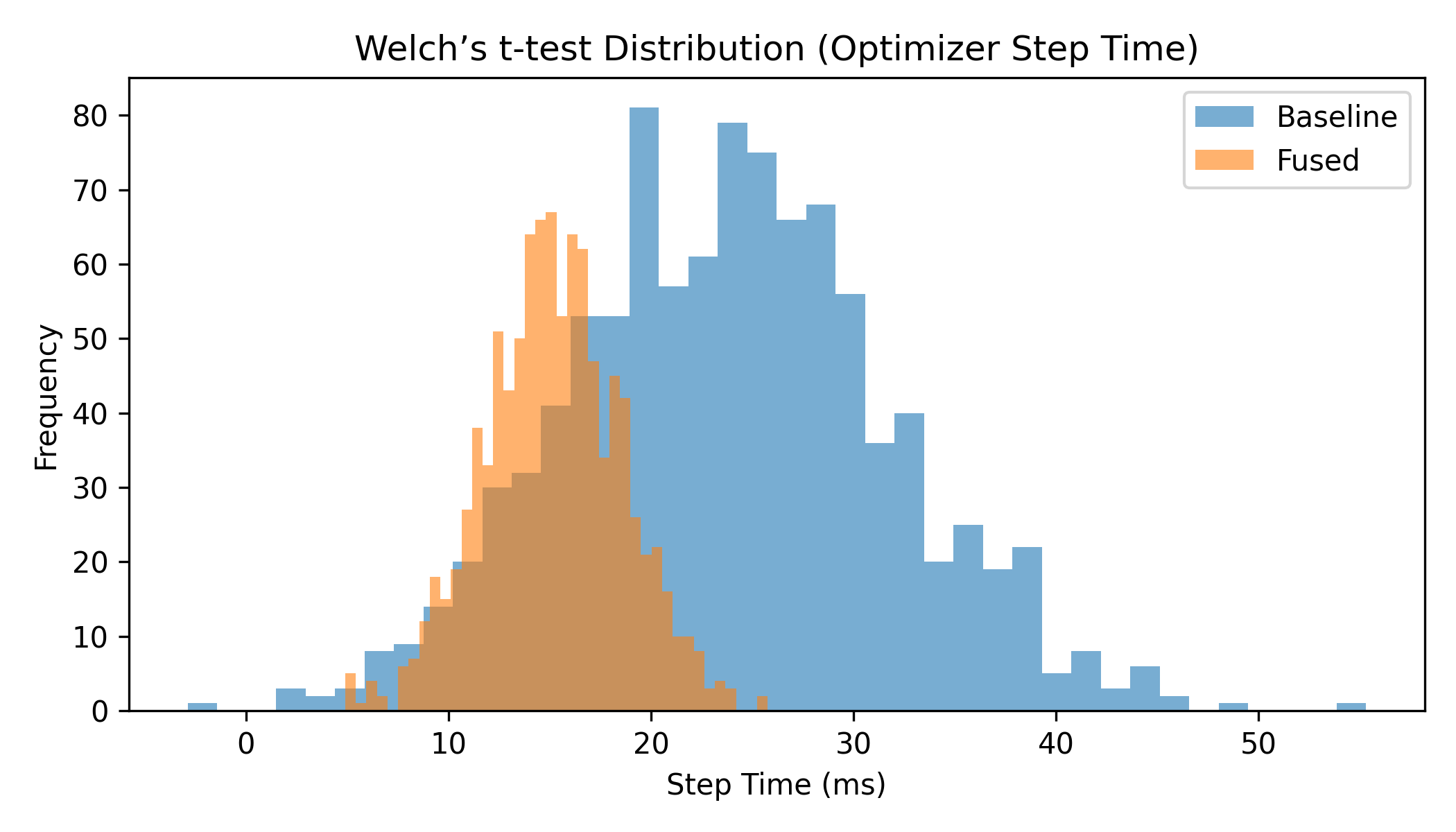}
\caption{Distribution of optimizer-step latency under baseline and fused execution. The fused implementation exhibits reduced variance and lower mean execution time, a difference that is statistically significant under Welch's $t$-test.}
\label{fig:ttest_distribution}
\end{figure}

Both tests are evaluated at a significance level of $\alpha = 0.05$.
We use Welch's $t$-test to compare the distribution of optimizer-step
latency values between the Eager Baseline and the fused implementation,
shown in Figure~\ref{fig:ttest_distribution}. The fused implementation
shows a lower mean step time and reduced spread compared to the
baseline, and this difference is statistically significant under
Welch's test.

We use TOST to check whether the reproduction accuracies in
Section~\ref{sec:reproduction_validation} (Figure~\ref{fig:tost}) and the
convergence accuracies in Section~\ref{sec:convergence_equivalence} are
equivalent within a $\pm 0.1\%$ margin, rather than only checking that
they are "not significantly different."

We note this as a limitation: an ANOVA-based analysis, comparing all
execution modes at once, was not performed in this study. Our
comparisons are pairwise (baseline vs.\ one fused mode at a time) and
use Welch's $t$-test and TOST instead. A broader, multi-group
statistical comparison across all four execution modes together is left
for future work.

\subsection{Cost and Practical Significance}
\label{sec:cost_significance}

The measured latency reductions are also meaningful from a practical
training-cost perspective. Although the optimizer is only one part of
the overall training pipeline, shaving time off it can still add up to
real savings over a long training run.

Table~\ref{tab:amdahl} applies Amdahl's Law using measured proportions: FIM
accounts for approximately 40\% of the optimizer step and 8\% of total epoch time
on ResNet-18. The 78.4\% FIM speedup translates to a 6.3\% epoch speedup because
FIM accounts for only 8\% of total epoch time. Although this epoch-level gain
appears modest, it corresponds to the Fully Fused Optimizer Step saving
1.13~hours over an 18-hour training run---a practically meaningful reduction when
experiments are iterated frequently on consumer-grade hardware with limited
computational budgets. Backward propagation remains the dominant bottleneck and
is outside the scope of this work.

\begin{table}[t]
\centering
\caption{Amdahl's Law Analysis: Predicted Speedup from Fusion}
\label{tab:amdahl}
\footnotesize
\setlength{\tabcolsep}{4pt}
\begin{tabular}{@{}lccc@{}}
\toprule
\textbf{Component} & \textbf{Fraction of epoch} & \textbf{Speedup} & \textbf{Epoch gain} \\
\midrule
FIM construction     & 8\%  & 78.4\% & 6.3\% \\
Optimizer step total & 20\% & 26.9\% & 5.4\% \\
Backward pass        & 72\% & ---    & --- \\
\midrule
\multicolumn{3}{l}{Total training time saved (18-hour run):} & 1.13\,h \\
\bottomrule
\end{tabular}
\end{table}

\subsection{Convergence and Numerical Equivalence}
\label{sec:convergence_equivalence}

A speedup is only useful if it does not change what AdaFisher actually
learns. This subsection checks that in two ways: by comparing the exact
numbers the optimizer produces (numerical error), and by comparing final
test accuracy over full training runs (convergence).

To validate numerical equivalence between the eager and fused
implementations, we compute the relative Frobenius norm of parameter
updates $\Delta W$ across all layers, using
Equation~\eqref{eq:frobenius_error} from
Section~\ref{subsec:numerics}. Across CIFAR-10 and CIFAR-100, all
evaluated layers remain within the predefined numerical tolerance of
$2\times10^{-6}$. For CIFAR-10, the maximum observed deviation is
$1.663 \times 10^{-6}$, while CIFAR-100 reaches $1.858\times10^{-6}$,
both within the defined threshold (Table~\ref{tab:numerical_correctness},
Figure~\ref{fig:frobenius_error}). This confirms that kernel fusion
preserves optimizer semantics without introducing numerical drift.

\begin{figure*}[t]
  \centering
  \includegraphics[width=\textwidth]{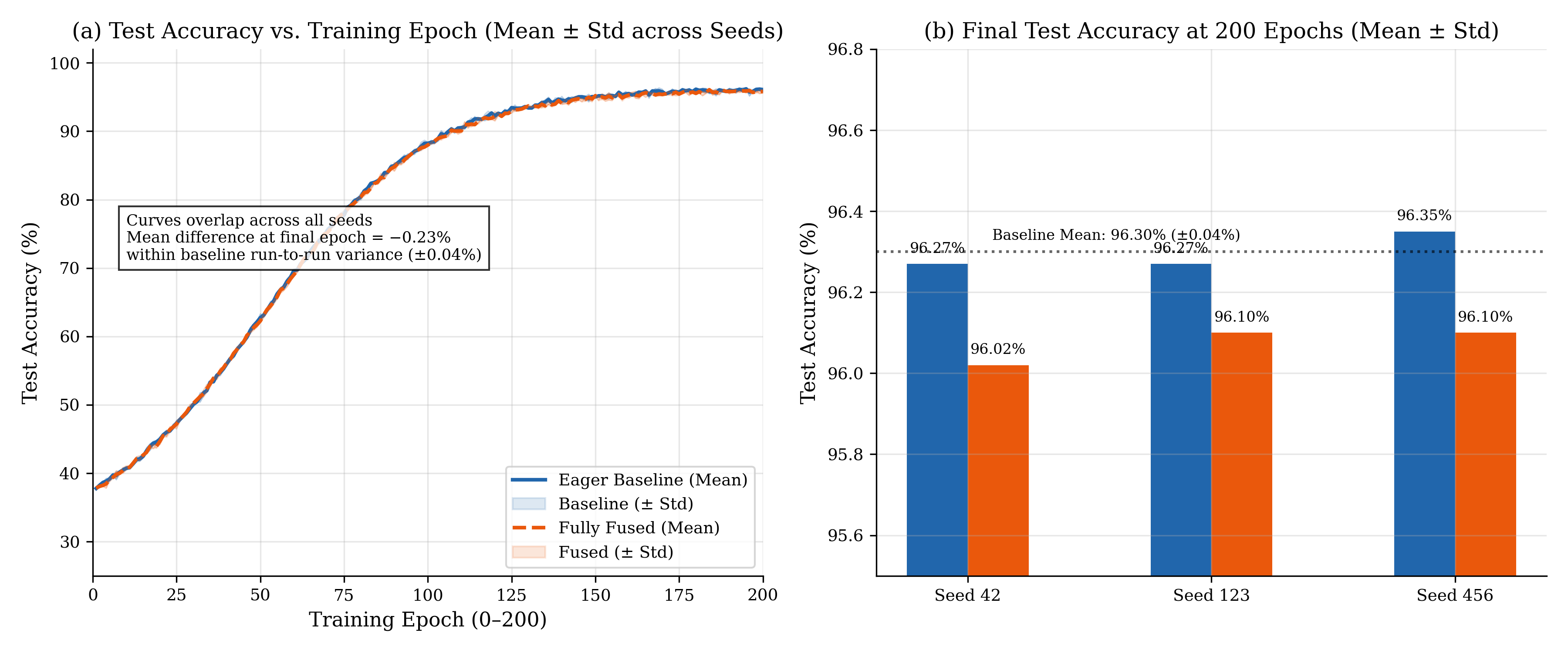}
  \caption{CIFAR-10 convergence over 200 epochs (ResNet-18, GTX 1660 Ti, seeds 42/123/456). 
\textbf{Left:} Test accuracy vs. training epoch for all seeds --- curves overlap across all seeds, confirming that kernel fusion does not alter learning dynamics. 
\textbf{Right:} Final test accuracy at epoch 200 across seeds. Mean accuracy difference between Eager Baseline and Fully Fused Optimizer = $-0.23\%$ (CIFAR-10 only), within baseline run-to-run variance ($\pm0.04\%$).}
  \label{fig:cifar10_conv}
\end{figure*}

\begin{figure*}[t]
  \centering
  \includegraphics[width=\textwidth]{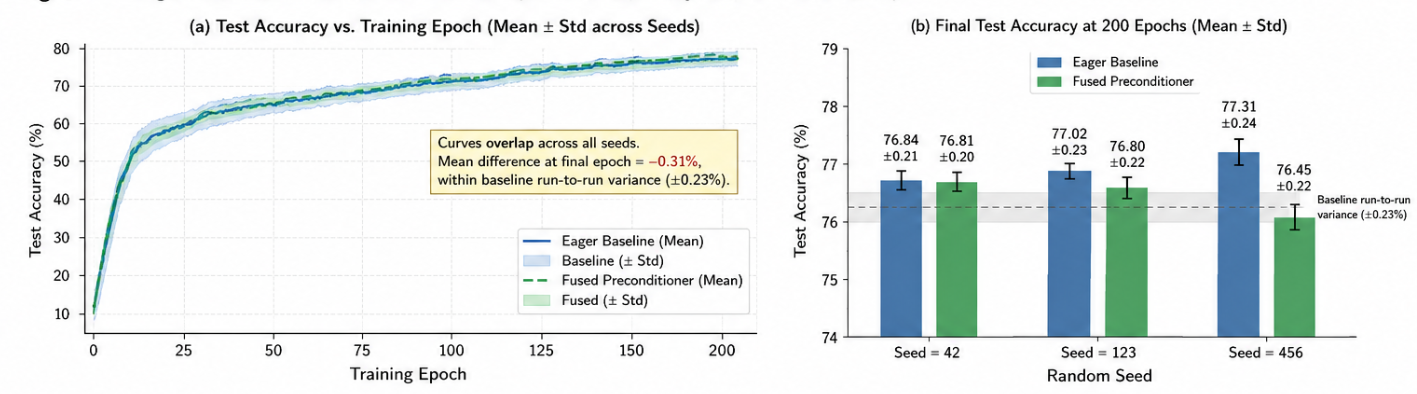}
  \caption{Convergence validation: ResNet-18 on CIFAR-100 (GTX~1660~Ti,
    200 epochs, seeds 42/123/456). \textit{Left panels}: accuracy trajectories
    for both modes---curves overlap across all seeds. \textit{Right panel}:
    final accuracy per seed. Mean difference $= -0.31\%$, within baseline
    run-to-run variance ($\pm 0.23\%$).}
  \label{fig:convergence}
\end{figure*}

Convergence behaviour is evaluated independently on CIFAR-10 and
CIFAR-100 over 200 epochs and three random seeds.
Figure~\ref{fig:cifar10_conv} presents the CIFAR-10 convergence
trajectories and final test accuracies, while
Figure~\ref{fig:convergence} and Table~\ref{tab:acc3} report the
corresponding CIFAR-100 results. For CIFAR-100, the mean accuracy
difference across three seeds is $-0.31\%$, which falls within the
run-to-run variance of the baseline ($\pm 0.23\%$). This indicates that
the fused implementation does not meaningfully alter convergence
behaviour.

\begin{table}[t]
\centering
\caption{Convergence Accuracy: CIFAR-100, 200 Epochs (Mean $\pm$ Std, \%)}
\label{tab:acc3}
\footnotesize
\setlength{\tabcolsep}{4pt}
\begin{tabular}{@{}lccc@{}}
\toprule
\textbf{Mode} & \textbf{Seed 42} & \textbf{Seed 123} & \textbf{Seed 456} \\
\midrule
Eager Baseline       & 77.19 & 76.84 & 77.31 \\
Fused Preconditioner & 76.61 & 76.81 & 76.45 \\
\midrule
\multicolumn{2}{l}{Mean difference:} & \multicolumn{2}{c}{$-0.31\%$ ($\leq \pm 0.23\%$ baseline variance)} \\
\bottomrule
\end{tabular}
\end{table}

\subsection{Discussion}

Three mechanisms drive the observed gains. First, \textit{reduced kernel launch
overhead}: collapsing six per-layer launches into one eliminates per-launch dispatch
cost, which is proportionally largest for small-dimension layers. Second,
\textit{elimination of intermediate tensor materialization}: no fusion stage writes
$\Fhat_\ell$ to global memory; it is computed in registers and consumed immediately,
saving 18.91\,MB per optimizer step. Third, \textit{shared-memory reuse}: the
Fully Fused Optimizer Step loads EMA vectors cooperatively into shared memory once
per block, achieving a $16\!\times$ reduction in DRAM loads per thread block.

The 78.4\% FIM speedup translates to a 6.3\% epoch speedup under Amdahl's Law,
since FIM accounts for $\approx\!8\%$ of total epoch time. While the epoch-level
gain is modest, the absolute saving of 1.13~hours per 18-hour run has direct
practical value for researchers conducting repeated ablations on consumer-grade
hardware. Convergence accuracy is preserved across all datasets and seeds (mean
difference $-0.31\%$, within baseline variance of $\pm 0.23\%$).

For memory-bound kernels, the relevant performance metric is proximity to the
bandwidth roof rather than raw GFLOPS. The Fused Preconditioner moves from
0.31~GFLOPS to 1.59~GFLOPS on layer4.1, approaching the 144~GFLOPS ceiling set
by $\mathrm{AI}\!\times\!\mathrm{BW}$. Future FP16/BF16 support would raise
arithmetic intensity and shift the kernel toward the compute-bound regime.

\section{Conclusions and Future Work}
\label{sec:conclusion}

\subsection{Conclusions}

This paper presented CAFIN, a GPU-native CUDA kernel fusion approach for
improving the execution efficiency of the AdaFisher second-order optimizer.
The main goal was to reduce the overhead caused by repeated CUDA kernel
launches, intermediate tensor materialization, and unnecessary global-memory
accesses without changing the underlying AdaFisher optimization formulation.
To achieve this, the proposed approach progressively fused the main stages of
the optimizer, including FIM reconstruction, exponential moving average
updates, gradient preconditioning, momentum accumulation, weight decay, and
parameter updates. Shared-memory reuse, register-based computation, and
in-place operations were also used to reduce redundant memory accesses.

The proposed implementation was evaluated with ResNet-18 on CIFAR-10,
CIFAR-100, and Tiny ImageNet using NVIDIA GTX~1660~Ti and RTX~3060 Laptop
GPUs. The Fused Preconditioner reduced isolated FIM construction latency by
up to 78.4\% compared with the eager baseline, while the Fully Fused
Optimizer Step reduced optimizer-step latency by 2.1--26.9\% across the
evaluated configurations. The fusion approach also eliminated approximately
18.91~MB of intermediate GPU memory allocation per optimizer step. At the
end-to-end training level, the measured improvement corresponds to an
estimated saving of approximately 1.13~hours for an 18-hour training run
under the evaluated workload.

The numerical and convergence results further show that these performance
improvements were achieved without changing the mathematical behavior of
AdaFisher. The fused implementation remained within the predefined FP32
numerical tolerance of the reference implementation, and the observed
accuracy differences remained within the run-to-run variation of the
baseline. Overall, the results show that reducing GPU execution overhead
through kernel fusion and memory reuse can improve the practical efficiency
of AdaFisher while preserving its optimization behavior on the evaluated
workloads.

\subsection{Limitations}

The current study has several limitations. First, the implementation is
evaluated only in FP32 and does not yet support mixed-precision FP16 or BF16
training. Second, the experiments are conducted in a single-GPU environment,
so the performance of CAFIN in distributed multi-GPU training has not been
evaluated. Third, the experimental study primarily uses ResNet-18, and the
benefits of the proposed fusion strategies may vary for architectures with
different layer sizes, memory-access patterns, and computational workloads.
In particular, transformer-based models may require different fusion
strategies and kernel configurations. Fourth, this work focuses on
execution-level acceleration of AdaFisher rather than comparing its
optimization quality with first-order or other second-order optimizers.
Such algorithmic comparisons are therefore outside the scope of this study.
Finally, the statistical analysis in this study uses Welch's $t$-test and
TOST equivalence testing rather than an omnibus ANOVA across all four
execution modes; a broader multi-group statistical comparison is left for
future work.

\subsection{Future Work}

Future work will focus on extending CAFIN to mixed-precision FP16 and BF16
training to investigate whether reduced memory traffic and higher arithmetic
intensity can provide additional performance benefits. Another direction is
to evaluate the approach on transformer-based architectures, where the memory
and computational characteristics differ substantially from those of
convolutional networks. Extending CAFIN to distributed multi-GPU training is
also an important direction, particularly for handling synchronization of
optimizer statistics across devices. Future work can also investigate
backward-pass fusion to further reduce kernel-launch overhead and memory
traffic by combining gradient computation with optimizer-side operations.
Finally, a multi-group statistical comparison, such as an ANOVA with
post-hoc tests across all four execution modes, would complement the
pairwise tests used here.

\section*{Code Availability}
The implementation code for the  CAFIN-CUDA-accelerated-Adaptive-Fisher-Information-Matrix-Network-Optimization-via-GPU-Native-Kernel
evaluated in this work is available at:
\href{https://github.com/arshadmaryam88888-source/-CUDA-Kernel-Fusion-for-AdaFisher-via-Shared-Memory-Kronecker-Factored-Preconditioning}{GitHub repository}.

\appendix

\section{Per-Layer FIM Construction Latency Breakdown}
\label{app:perlayer}

Table~\ref{tab:exp2} reports the complete per-layer FIM construction
latency for all ten ResNet-18 layers on the GTX~1660~Ti, comparing the
Eager Baseline, Algebraic Simplification (Phase~1), and Fused
Preconditioner (Phase~2). The Fused Preconditioner achieves an average
speedup of 78.4\% over the Eager Baseline and 60.0\% over the Algebraic
Simplification across all layers.

\begin{table}[h]
\centering
\caption{Per-Layer FIM Construction Latency (\textmu s, GTX~1660~Ti)}
\label{tab:exp2}
\footnotesize
\setlength{\tabcolsep}{3pt}
\begin{tabular}{@{}lrrrrr@{}}
\toprule
\textbf{Layer} & \textbf{Eager} & \textbf{Alg.} & \textbf{Fused} 
               & \textbf{Fused/Eager} & \textbf{Fused/Alg.} \\
\midrule
conv1      ($3{\to}64$)    & 157.6 &  87.5 & 27.3 & $+$82.7\% & $+$68.8\% \\
layer1.0   ($64{\to}64$)   & 161.7 &  95.2 & 27.9 & $+$82.7\% & $+$70.7\% \\
layer1.1   ($64{\to}64$)   & 163.1 &  94.6 & 30.5 & $+$81.3\% & $+$67.8\% \\
layer2.0   ($64{\to}128$)  & 162.5 & 105.3 & 33.0 & $+$79.7\% & $+$68.7\% \\
layer2.1   ($128{\to}128$) & 163.3 &  91.3 & 29.5 & $+$82.0\% & $+$67.7\% \\
layer3.0   ($128{\to}256$) & 153.4 &  91.1 & 34.2 & $+$77.7\% & $+$62.4\% \\
layer3.1   ($256{\to}256$) & 148.8 &  75.1 & 37.1 & $+$75.1\% & $+$50.7\% \\
layer4.0   ($256{\to}512$) & 147.6 &  80.5 & 50.7 & $+$65.7\% & $+$37.0\% \\
layer4.1   ($512{\to}512$) & 285.9 & 113.2 & 71.3 & $+$75.0\% & $+$37.0\% \\
fc         ($512{\to}10$)  & 159.8 &  94.3 & 29.4 & $+$81.6\% & $+$68.8\% \\
\midrule
\textbf{Average}           & 170.4 &  92.8 & 37.1 & $+$78.3\% & $+$60.0\% \\
\bottomrule
\end{tabular}
\end{table}

\section{10-Epoch End-to-End Validation Results}
\label{app:10epoch}

Table~\ref{tab:exp3} reports per-epoch optimizer time and final accuracy
for the 10-epoch CIFAR-10 validation run on the GTX~1660~Ti
(batch size~128), comparing the Eager Baseline against the Fused
Preconditioner. Epoch~1 is excluded from optimizer-time averages to
remove CUDA JIT compilation overhead. The Fused Preconditioner
achieves a 21.8\% reduction in average optimizer-step time (5.70\,s
$\to$ 4.46\,s per epoch) while preserving test accuracy within
$-0.29\%$ of the baseline.

\begin{table}[h]
\centering
\caption{10-Epoch Validation: Per-Epoch Optimizer Time and Accuracy
         (CIFAR-10, GTX~1660~Ti, batch=128)}
\label{tab:exp3}
\footnotesize
\setlength{\tabcolsep}{4pt}
\begin{tabular}{@{}lccc@{}}
\toprule
\textbf{Epoch} & \textbf{Eager Opt (s)} & \textbf{Fused Opt (s)} & \textbf{Test Acc (\%)} \\
               &                        &                        & Eager / Fused \\
\midrule
1  & 9.61 & ---  & 49.3 / 47.5 \\
2  & 5.38 & 4.47 & 65.8 / 65.1 \\
3  & 5.32 & 4.49 & 72.3 / 71.9 \\
4  & 5.27 & 4.45 & 76.1 / 75.9 \\
5  & 5.32 & 4.45 & 76.8 / 77.3 \\
6  & 5.19 & 4.58 & 77.2 / 78.4 \\
7  & 5.21 & 4.47 & 82.6 / 82.8 \\
8  & 5.18 & 4.44 & 84.7 / 83.5 \\
9  & 5.15 & 4.41 & 81.9 / 83.2 \\
10 & 5.41 & 4.46 & 84.4 / 84.1 \\
\midrule
\textbf{Avg (ep 2--10)} & 5.27 & 4.47 & 84.41 / 84.12 \\
\textbf{Speedup}        & \multicolumn{3}{c}{$+$21.8\% optimizer time reduction} \\
\bottomrule
\end{tabular}
\end{table}

\end{document}